\documentclass[lettersize,journal]{IEEEtran}
\usepackage{times,latexsym}
\usepackage[T1]{fontenc}

\usepackage[linesnumbered,ruled,noend]{algorithm2e}
\usepackage{algpseudocode}

\usepackage{amssymb}
\usepackage{booktabs}
\usepackage{multirow}
\usepackage{multicol}
\usepackage{makecell}

\usepackage{amsmath,amsfonts}
\usepackage{mathtools}
\usepackage{amsthm}
\usepackage{array}
\usepackage[caption=false,font=normalsize,labelfont=sf,textfont=sf]{subfig}
\usepackage{textcomp}
\usepackage{stfloats}
\usepackage{url}
\usepackage{verbatim}
\usepackage{graphicx}
\usepackage{cite}

\usepackage{threeparttablex}

\usepackage{hyperref}
\usepackage{cleveref}

\DeclareMathOperator*{\argmax}{arg\,max}

\begin{document}

\title{Beyond Elicitation: Provision-based Prompt Optimization\\ for Knowledge-Intensive Tasks}

\author{Yunzhe Xu,
        Zhuosheng Zhang,
        Zhe Liu
\thanks{This paper was supported by the National Natural Science Foundation of China under Grant 62303307, and in part by the National Key Laboratory of Human Machine Hybrid Augmented Intelligence, Xi’an Jiaotong University (No. HMHAI-202408). (Corresponding author: Zhe Liu.)}
\thanks{Yunzhe Xu is with the School of Computer Science, Shanghai Jiao Tong University, Shanghai 200240, China (e-mail: xyz9911@sjtu.edu.cn). Zhuosheng Zhang is with the School of Computer Science, Shanghai Jiao Tong University, Shanghai 200240, China (e-mail: zhangzs@sjtu.edu.cn). Zhe Liu is with the School of Automation and Intelligent Sensing, Shanghai Jiao Tong University, Shanghai 200240, China. Zhe Liu is also with the National Key Laboratory of Human-Machine Hybrid Augmented Intelligence, Institute of Artificial Intelligence and Robotics, Xi’an Jiaotong University, Xi’an 710049, China. (e-mail: liuzhesjtu@sjtu.edu.cn).}
\thanks{Code is available at \href{https://github.com/xyz9911/KPPO}{https://github.com/xyz9911/KPPO}.}
}



\maketitle

\begin{abstract}
While prompt optimization has emerged as a critical technique for enhancing language model performance, existing approaches primarily focus on elicitation-based strategies that search for optimal prompts to activate models' capabilities. These methods exhibit fundamental limitations when addressing knowledge-intensive tasks, as they operate within static knowledge capacity rather than providing the factual knowledge, terminology precision, and reasoning patterns required in specialized domains. To address these limitations, we propose Knowledge-Provision-based Prompt Optimization (KPPO), a framework that reformulates prompt optimization as systematic knowledge integration rather than potential elicitation. KPPO introduces three key innovations: 1) a knowledge gap filling mechanism for knowledge gap identification and targeted remediation; 2) a batch-wise candidate evaluation approach that considers both performance improvement and distributional stability; 3) an adaptive knowledge pruning strategy that balances performance and token efficiency, reducing up to 29\% of inference token usage. Evaluation on 15 knowledge-intensive benchmarks from various domains demonstrates KPPO's superiority over elicitation-based methods, with an average improvement of \textasciitilde6\% over baselines while achieving comparable or lower token consumption.
\end{abstract}


\section{Introduction}

\IEEEPARstart{L}{arge} Language Models (LLMs) have achieved unprecedented performance \cite{zhang2025igniting} across diverse natural language processing tasks through sophisticated prompt engineering techniques~\cite{brown2020gpt3}. The field has evolved from manual prompt design approaches~\cite{wei2022cot,zhou2023leasttomost} to automated optimization frameworks~\cite{pryzant2023apo,yang2024opro,wang2024promptagent,yuksekgonul2024textgrad}, where optimizer LLMs iteratively refine prompts to maximize task performance. These automated approaches, collectively termed \textit{elicitation-based optimization}, operate under the fundamental assumption that optimal prompts can unlock latent capabilities within pre-trained model parameters through strategic reformulation of instructions, exemplars, or reasoning templates. However, elicitation-based optimization encounters fundamental limitations when applied to knowledge-intensive domains that require specialized expertise beyond the model's parametric knowledge. Consider tasks in specialized scientific domains, emerging technologies, or domain-specific applications where factual accuracy, terminology precision, and specialized reasoning patterns are paramount, the assumption that optimal prompts can elicit non-existent knowledge becomes untenable.

\begin{figure}[t]
\centering
\includegraphics[scale=0.5]{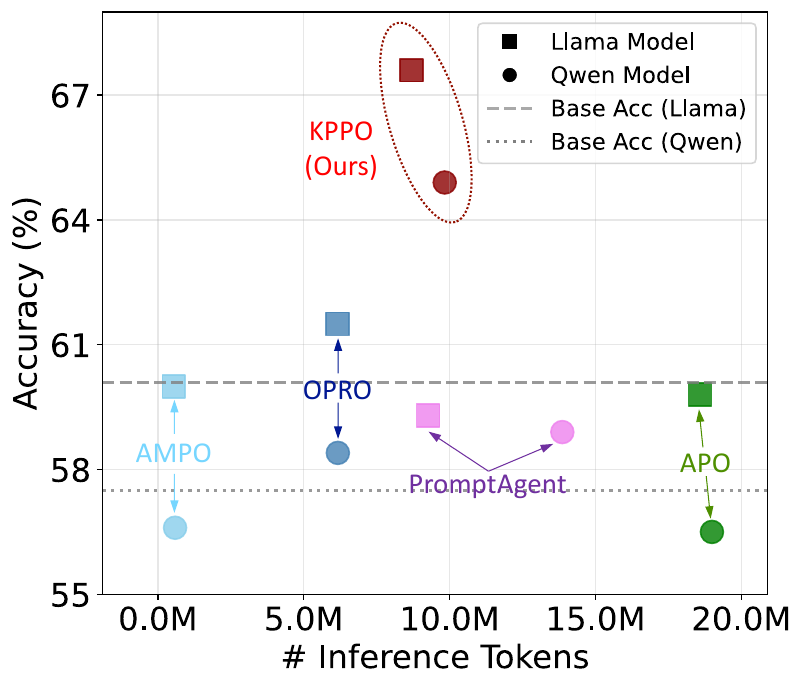}
\caption{Analysis of prompt optimization on 15 knowledge-intensive tasks from various domains. Traditional elicitation-based methods achieve marginal or even negative improvements, while KPPO demonstrates substantial improvements (average +6\%) while achieving comparable or enhanced efficiency.
}
\label{fig:intro}
\end{figure}

\begin{figure*}[t]
\centering
\includegraphics[scale=0.48]{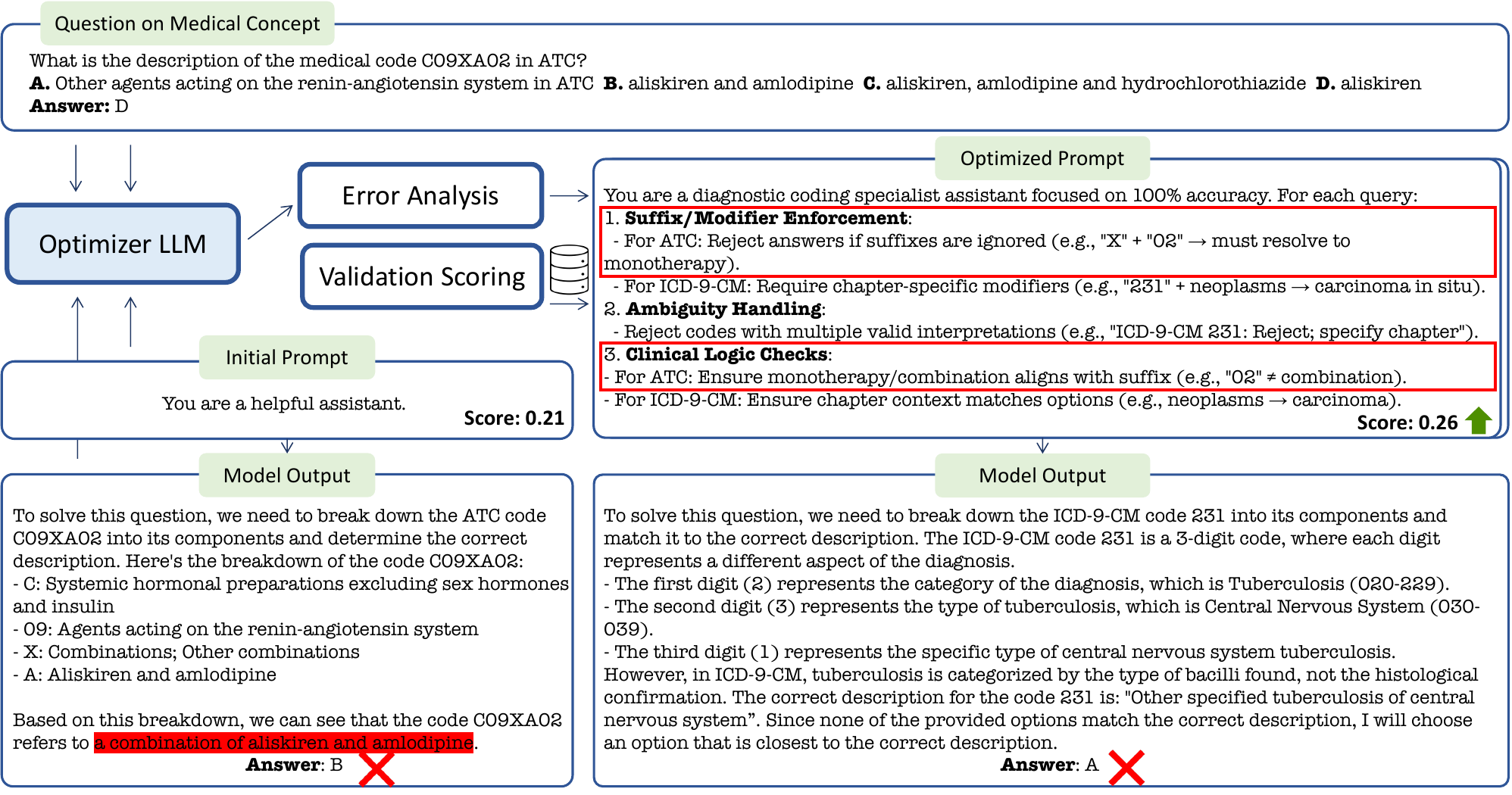}
\caption{Visualization of the failure of elicitation-based prompt optimization to fill LLM's knowledge gap in specific domains. Despite achieving improved accuracy on the validation split, the optimized prompt fails to provide sufficient domain knowledge, resulting in continued errors on the original failure cases. The optimized prompts capture surface-level patterns rather than providing the substantive domain knowledge required to resolve the failure cases.
}
\label{fig:pilot_1}
\end{figure*}


We analyze the performance of traditional methods across 15 knowledge-intensive question-answering benchmarks spanning multiple domains. As demonstrated in \Cref{fig:intro}, traditional methods achieve marginal or even negative improvements when a clear task instruction is provided. \Cref{fig:pilot_1} reveals that elicitation-based optimization fails to address underlying knowledge gaps, instead producing shallow improvements that do not resolve the fundamental knowledge deficiencies. Our systematic analysis of this phenomenon reveals two critical deficiencies: \textbf{1) Knowledge Poverty}: optimized prompts lack sufficient domain-specific information to address knowledge gaps, instead focusing on superficial reformulations that fail to provide substantive content. \textbf{2) Optimization Ineffectiveness}: even when prompts incorporate relevant information, the optimization process lacks effective validation mechanisms to ensure the knowledge gap is mitigated.

To address these fundamental deficiencies, we must shift from elicitation to provision by directly integrating domain-specific knowledge rather than attempting to activate non-existent parametric knowledge. However, this introduces three challenges: How can we identify missing domain knowledge and generate targeted prompts to fill these gaps? How can we ensure selected prompts reliably address knowledge gaps rather than achieving spurious validation scores? How can we balance comprehensive knowledge provision with token efficiency constraints as knowledge accumulates iteratively?

We propose \textbf{Knowledge-Provision-based Prompt Optimization (KPPO)}, a principled framework that addresses these challenges through three corresponding innovations. Rather than attempting to elicit non-existent knowledge through linguistic manipulation, KPPO explicitly provides the factual content, terminology, and reasoning patterns required for task success. First, an analysis framework examines failure cases to identify specific knowledge deficiencies and generates candidate prompts incorporating targeted domain content. Second, we introduce batch-wise dual-objective evaluation that jointly optimizes for performance improvement on recent failures and distributional stability, inspired by trust region methods~\cite{trpo}. Third, we design an adaptive pruning mechanism that identifies structural inefficiencies through local degree and global balance constraints, maintaining essential domain information while reducing inference token consumption.

Evaluation across 15 knowledge-intensive benchmarks demonstrates KPPO's effectiveness. Our method achieves average improvements of 6.1\% and 6.0\% on Llama 3.1 \cite{dubey2024llama3.1} and Qwen 2.5 \cite{qwen2025qwen2.5} respectively. Notably, the pruning variant significantly reduces token usage by up to 29\% while outperforming baselines. These results establish a new direction for prompt optimization research, extending beyond the knowledge capacity of pre-trained models to systematically incorporate external knowledge. This innovation has significant implications for deploying LLMs in specialized domains where tuning approaches are computationally prohibitive or data-limited. The contributions of this work are threefold:

\begin{itemize}
    \item We propose KPPO, a framework that systematically integrates domain-specific knowledge into prompts, achieving significant improvements across knowledge-intensive tasks spanning various domains.
    
    \item The technical innovations include: knowledge gap filling with targeted gap identification, dual-objective batch-wise evaluation ensuring both performance gains and optimization robustness, and adaptive pruning that maintains effectiveness while reducing inference token consumption.
    
    \item We establish provision-based optimization as a viable paradigm for knowledge-intensive applications, addressing knowledge poverty and optimization ineffectiveness and providing an alternative for domain-specific tasks.
\end{itemize}

\section{Related Work}

\subsection{Prompt Optimization}
Prompt engineering has emerged as a critical technique for enhancing LLM performance, with research spanning multiple optimization dimensions including prompt design~\cite{wei2022cot,zhou2023leasttomost}, exemplar selection~\cite{zhang2023autocot,sorensen2022informationtheoretic}, and exemplar ordering~\cite{lu2022fantastically}. Early approaches relied on manual prompt crafting~\cite{zhao2021calibrate}, but recent work has shifted toward automated optimization to reduce engineering effort. One direction focuses on test-time prompt modification~\cite{cheng2023bpo,zhang2023tempera,kong2024prewrite,sun2024querydependent}, which refines user inputs to improve task understanding, either through auxiliary models~\cite{lin2023instinct,kwon2024stableprompt,chen2024instructzero} trained for prompt rewriting or through direct optimization. In contrast, automatic prompt optimization operates without auxiliary models during inference, employing only an optimizer LLM~\cite{zhou2022ape,yang2024opro,ye2023pe2,yang2024ampo} for iterative prompt refinement, offering greater practical efficiency. Current approaches primarily focus on comprehensive search algorithms, employing beam search~\cite{pryzant2023apo,yang2024ampo}, Monte Carlo tree search~\cite{wang2024promptagent}, and evolutionary algorithms~\cite{guo2024evoprompt,hazman2025grammar} to generate diverse mutations~\cite{prasad2023grips,fernando2024promptbreeder} with bandit-based selection on validation sets~\cite{shi2024triple}. These methods collectively constitute elicitation-based optimization, seeking optimal prompts to unlock latent model capabilities. Recent extensions have broadened the scope to multi-turn scenarios~\cite{chen2024promst}, label-free optimization~\cite{zhang2024glape}, exemplar-focused refinement~\cite{long2024advicl,agarwal2024promptwizard,cui2025see}, and advanced techniques incorporating multi-agent learning~\cite{zhang2025mars}, meta-learning~\cite{choi2025system}, and probabilistic optimization~\cite{zhao2025pmpo}. While these approaches effectively improve prompt quality through strategic reformulation, they fundamentally operate within static knowledge capacity by seeking to elicit latent capabilities rather than augmenting the model's knowledge base with external domain expertise. Our work diverges from this elicitation paradigm by systematically integrating domain-specific knowledge.

\subsection{Self-Evolution in Large Language Models}
LLMs have demonstrated potential for self-improvement through mechanisms that identify and rectify errors without external supervision~\cite{pan2024survey,welleck2023selfcorrect}. Initial approaches focus on single-turn refinement through self-reflection techniques~\cite{madaan2024selfrefine,ling2024deductive}, where models analyze their own reasoning traces to detect inconsistencies and improve response quality. This verification process can be augmented by invoking external tools~\cite{gou2024critic,zhao2023verifyedit} to ensure factual consistency and logical coherence. More recent work extends self-evolution to multi-turn scenarios, learning from direct environmental feedback~\cite{shinn2024reflexion,zhang2024agentpro} to iteratively improve task performance through experience accumulation and reflection in sequential decision-making contexts for language agents. TextGrad~\cite{yuksekgonul2024textgrad} bridges self-evolution with prompt optimization by enabling textual gradients derived from self-reflection to backpropagate through prompts. However, when models lack domain-specific knowledge, self-reflection mechanisms cannot bridge this gap. In contrast, our approach directly addresses knowledge gaps by systematically integrating external domain content into prompts, transcending the knowledge boundaries that constrain self-evolution methods. While self-evolution improves how models use existing knowledge, KPPO expands what knowledge is available to the model during inference.

\section{Pilot Experiments}

This section provides diagnoses for elicitation-based prompt optimization that motivates our approach.

\begin{figure}[t]
\centering
\includegraphics[scale=0.3]{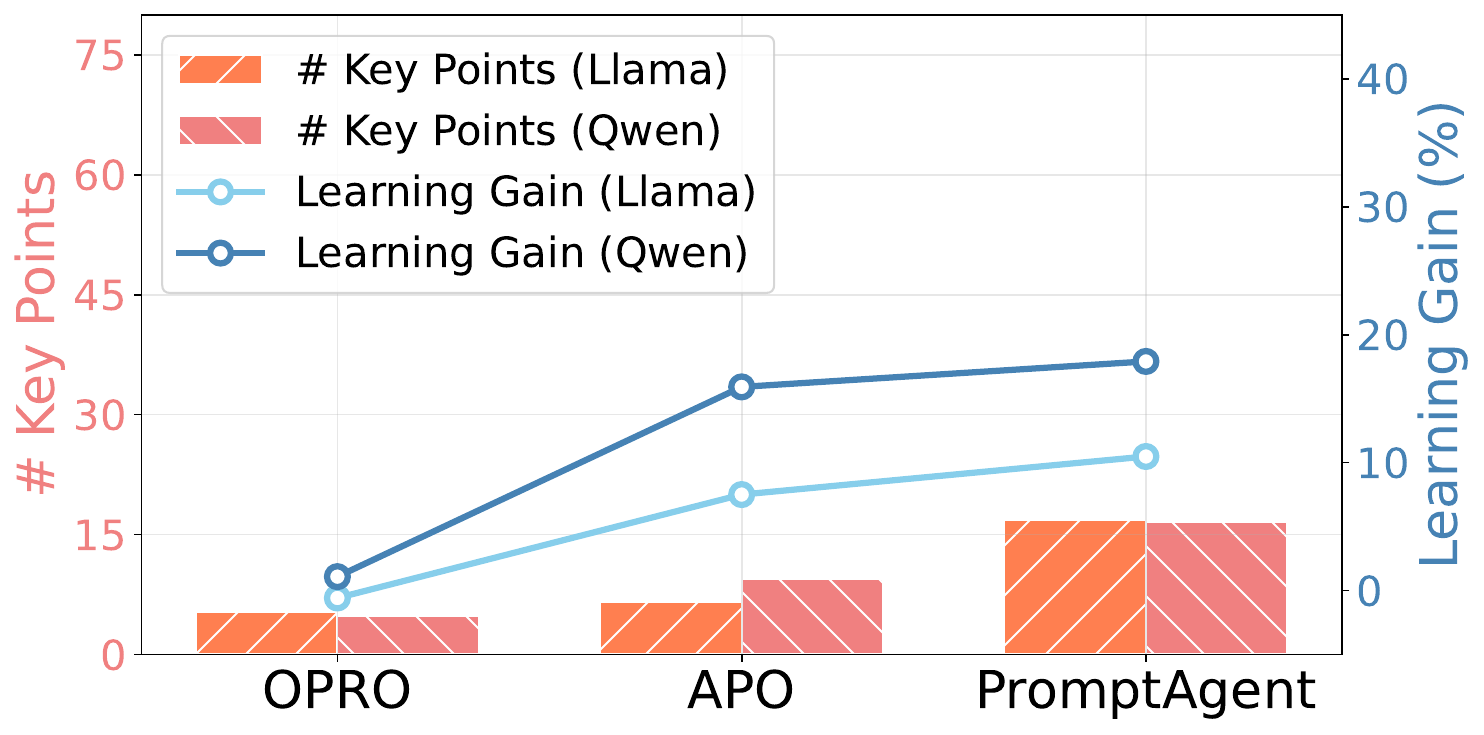}
\caption{Analysis of key points within the optimized prompt and learning gain across knowledge-intensive tasks.
}
\label{fig:pilot_2}
\end{figure}

\subsection{Problem Formulation}

Given a dataset $\mathcal{D}=\{(x_i, y_i)\}_{i=1}^N$ where each instance $(x_i, y_i)$ represents an input-output pair requiring specialized domain knowledge, we define a system prompt $p$ as a structured natural language construct containing both domain-specific knowledge and procedural guidance for task execution. The inference process combines the prompt $p$, task input $x$, and standardized task instruction $\ell$ to produce a prediction $\hat{y} = \text{LLM}_{\text{task}}(p, x, \ell)$ from the target language model. The optimization objective seeks to identify the optimal prompt $p^*$ that maximizes expected task performance:

\begin{equation}
    \begin{aligned}
        p^* &= \argmax_p \mathbb{E}_{(x,y)\sim\mathcal{D}}[f(p,x,y)],\\
        f(p,x,y) &= \begin{cases}
            1, & \text{if}\ \text{LLM}_{task}(p,x,\ell) = y \\
            0, & \text{otherwise},
        \end{cases}
    \end{aligned}
\end{equation}

\noindent where $f(\cdot)$ evaluates prediction correctness.

\subsection{Analysis of Elicitation-based Methods}

We evaluate three representative prompt optimization methods: OPRO \cite{yang2024opro}, APO \cite{pryzant2023apo}, and PromptAgent \cite{wang2024promptagent} on Llama 3.1 and Qwen 2.5. For each optimized prompt, we employ DeepSeek-V3 \cite{liu2024deepseek} as an evaluator to identify and quantify domain-specific key points through prompt analysis. We introduce the \textbf{learning gain} metric to quantify optimization effectiveness, capturing the model's ability to resolve previously failed cases through knowledge acquisition:

\small
\begin{equation}
\frac{1}{|P|} \sum_{i=1}^{|P|} \frac{1}{|\mathcal{B}_i|} \sum_{(x,y) \in \mathcal{B}_i} [f(p_i,x,y) - f(p_{i-1},x,y)],
\end{equation}
\normalsize

\noindent where $P$ represents the optimization trajectory and $\mathcal{B}_i$ is the training batch for step $i$. \Cref{fig:pilot_2} presents our empirical findings, revealing two critical deficiencies in current approaches:

\textbf{1) Knowledge Poverty}: All evaluated methods demonstrate remarkably low key points counts (<15), indicating optimization is not focusing on substantive knowledge integration.

\textbf{2) Optimization Ineffectiveness}: Learning gains remain substantially low (<20\%) across all methods, revealing that existing methods lack effective mechanisms to ensure that knowledge gaps have been effectively addressed. Prompts may achieve improved validation accuracy through pattern matching while failing to resolve underlying knowledge deficiencies.

These findings motivate our approach, which transcends elicitation-based limitations through systematic knowledge integration that directly fills knowledge gaps and improves optimization effectiveness.

\begin{figure*}[t]
  \centering
  \includegraphics[scale=0.52]{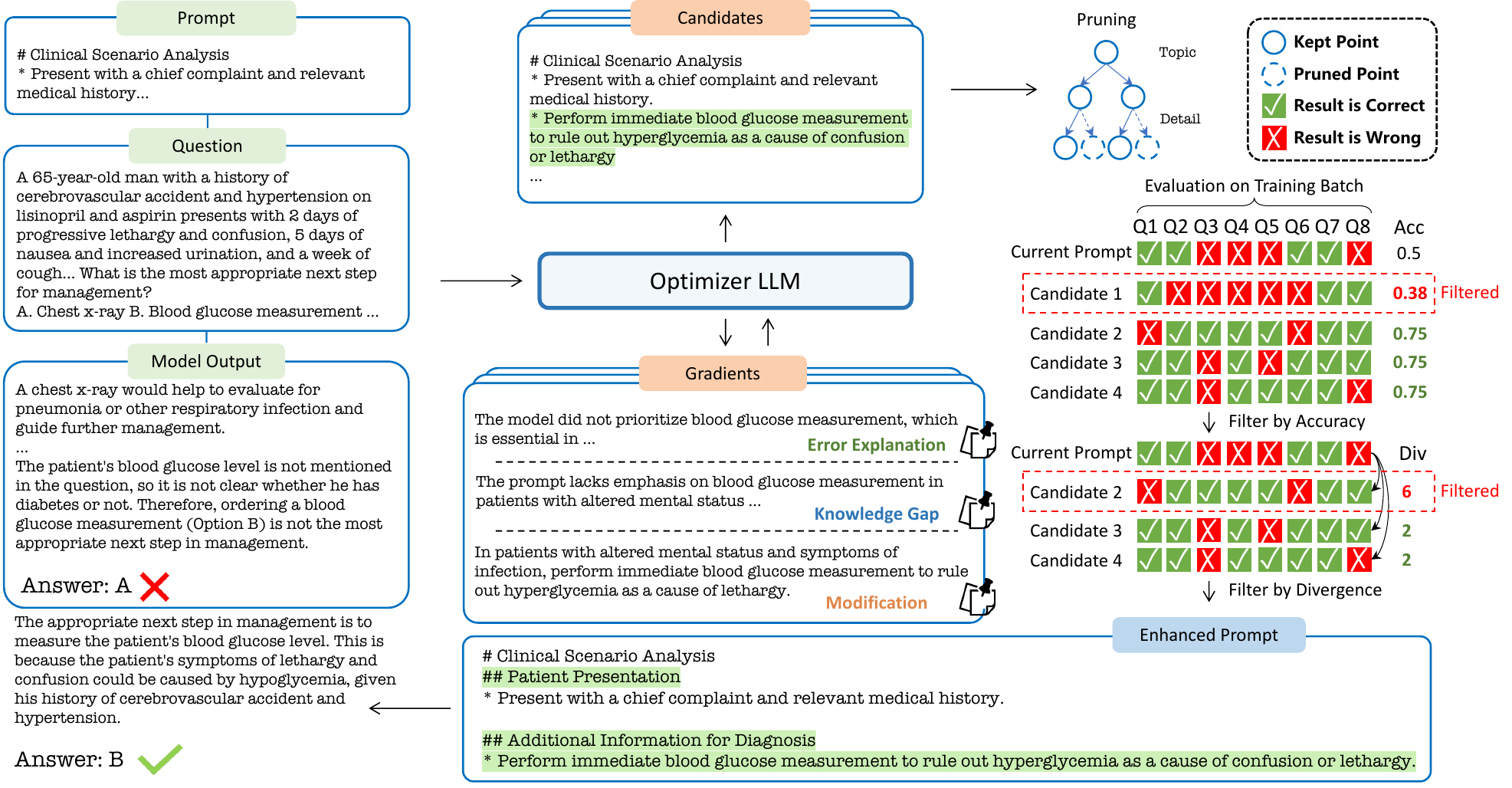}
  \caption{Overview of KPPO. Given task mistakes, the optimizer LLM generates ``gradients'' by analyzing the original prompt's limitations, producing explanations of failures, identifying knowledge gaps, and suggesting targeted modifications. The framework then integrates these gradients to generate candidate prompts, which undergo an alternative pruning procedure to avoid over-lengthened prompts. Prompt candidates are filtered with a batch-wise dual-objective evaluation that jointly considers performance improvement on recent training instances and distribution stability to ensure robust knowledge integration.}
  \label{fig:method}
\end{figure*}

\section{Method}
\subsection*{Overview}
KPPO addresses the limitations of elicitation-based approaches by integrating domain-specific knowledge into prompt structures. As illustrated in \Cref{fig:method}, KPPO employs an optimizer LLM ($\text{LLM}_{opt}$) to iteratively enhance prompts through targeted knowledge provision. The framework maintains a beam of optimized prompts $\mathcal{C}$ that are refined based on failure case analysis. At each iteration $t$, the system samples a batch $\mathcal{B}_t$ from the dataset, identifies failure cases $\mathcal{E}_t$, and generates improved prompts through analysis and candidate filtering. The optimization process is detailed in \Cref{alg:method}.

\begin{algorithm}[t]
\caption{KPPO}
\label{alg:method}
\small
\SetEndCharOfAlgoLine{}
\SetKwInOut{Input}{Input}
\SetKwInOut{Output}{Output}
\Input{
    Initial prompt $p_0$,
    Training set $\mathcal{D}_{train}$,
    Window size $K$,
    Iteration limit $T$,
    Validation set $\mathcal{D}_{val}$,
    New prompts per step $M$
}
\Output{Optimized knowledge prompt $p^*$}
\BlankLine
Initialize beam $C \leftarrow \{p_0\}$ and bank $\mathcal{Q} \leftarrow \emptyset$ \\
\For{step $t = 1$ \KwTo $T$}{
    Initialize candidate set $cand \leftarrow \emptyset$ \\
    Sample batch $\mathcal{B}_{t} \subset \mathcal{D}_{train}$ \\
    Update instance bank: $\mathcal{Q} \leftarrow \mathcal{Q} \cup \mathcal{B}_{t}$ \\
    Get $K$ recent instances: $\mathcal{Q}_K \leftarrow \{q_i \in \mathcal{Q} : i > |\mathcal{Q}| - K\}$ \\
    Get failure cases: $\mathcal{E} \leftarrow \{(x,y) \in \mathcal{B}_{t} : f(p,x,y)=0\}$ \\
    \For{$p \in C$}{
        $cand \leftarrow cand \cup \{(p,p)\}$ \\
        $issues \leftarrow \emptyset$ \\
        \If{pruning}{$issues \leftarrow \text{ViolationDetect}(p)$ \Comment{\Cref{alg:detection}} \\}
        \For{$i = 1$ \KwTo $M$}{
            Generate gradients: $g_i \leftarrow \text{LLM}_{opt}(p,\mathcal{E})$ \\
            Generate candidate: $p'_i \leftarrow \text{LLM}_{opt}(p,\mathcal{E},g_i,issues)$ \\
            $cand \leftarrow cand \cup \{(p'_i,p)\}$
        }
    }
    $C \leftarrow \text{CandidateEvaluate}(cand, \mathcal{Q}_K)$ \Comment{\Cref{alg:filter}} \\
}
\Return $\argmax_{p \in C} \mathbb{E}_{(x,y)\sim\mathcal{D}_{val}}[f(p,x,y)]$
\end{algorithm}

\subsection{Knowledge Gap Filling}
Our approach identifies knowledge deficiencies that cause prediction errors, moving beyond surface-level pattern recognition to address knowledge gaps. Given a current prompt $p$ and failure cases $\mathcal{E} = \{(x,y) \in \mathcal{B}_t : \text{LLM}_{task}(p,x,\ell) \neq y\}$, we employ $\text{LLM}_{opt}$ to analyze each failure from three perspectives: \textbf{1) Error Explanation}: Analyzes the underlying reasons why the current prompt fails, examining flaws in LLM's response. \textbf{2) Knowledge Gap Analysis}: Identifies specific domain knowledge missing from the current prompt that is required to resolve the failure cases. \textbf{3) Modification}: Suggests targeted improvements and specific knowledge pieces that should be integrated into the prompt structure. This analysis represents the ``gradients'' $g$ that provide guidance for prompt improvement. Based on these gradients $g$, $\text{LLM}_{opt}$ generates $M$ candidate prompts per iteration, each incorporating targeted knowledge to address identified deficiencies. To ensure systematic knowledge integration, KPPO organizes the incorporated domain information within each prompt using a structured knowledge hierarchy represented as a directed acyclic graph $\mathcal{T} = (V, E)$. The vertex set $V$ is partitioned into two disjoint subsets: \textbf{topic nodes} $V_t$, representing non-terminal conceptual categories, and \textbf{note nodes} $V_n$, representing strictly terminal leaf nodes containing atomic domain facts.

\begin{algorithm}[t]
\caption{Candidate Evaluation}
\label{alg:filter}
\small
\SetEndCharOfAlgoLine{}
\SetKwInOut{Input}{Input}
\SetKwInOut{Output}{Output}
\Input{
    Candidate pairs $cand$,
    Recent examples $\mathcal{Q}_K$,
    Beam width $W$
}
\Output{Selected prompts $B$}
\BlankLine
Initialize $ranking \leftarrow \emptyset$ \\
\For{$(p',p) \in \text{cand}$}{
    Calculate $\Delta s \leftarrow \Delta S(p',p)$ using \Cref{eq:gain} \\
    \If{$\Delta s > 0$}{
        Calculate $d \leftarrow D(p',p)$ using \Cref{eq:div} \\
        $ranking \leftarrow ranking \cup \{(\Delta s, d)\}$
    }
}
Sort $ranking$ by $(\Delta s, -d)$ \\
\Return Top $W$ prompts from $ranking$
\end{algorithm}

\subsection{Batch-Wise Candidate Evaluation}
The newly generated prompt candidates require systematic evaluation to filter out ineffective variants and ensure optimization robustness. Traditional prompt optimization relies on full validation set evaluation, which is computationally expensive and may not effectively validate knowledge integration quality. We introduce a batch-wise evaluation mechanism that operates on recent training instances $\mathcal{Q}_K$ to improve efficiency and robustness. However, accuracy-based batch-wise validation may lead to instability as the evaluation instances are limited to a fixed window $K$, and we cannot fully ensure that prompt changes will generalize well to other instances.

To address this challenge, we propose a principled filtering mechanism inspired by trust region policy optimization \cite{trpo} that considers both learning gains and output stability. Let $p(y|x)$ and $p'(y|x)$ denote output distributions under the current and candidate prompt respectively. We formulate the prompt update objective as:

\begin{equation}
\small
\begin{aligned}
    \argmax_{p'} \mathbb{E}_{(x,y)\sim\mathcal{Q}_K}\left[\frac{p'(y|x)}{p(y|x)}A(p',p,x,y)\right],
\end{aligned}
\end{equation}

\noindent where the ratio $\frac{p'(y|x)}{p(y|x)}$ measures distribution change and $A(p',p,x,y)=f(p',x,y) - f(p,x,y)$ represents the advantage of the new prompt over the current one. We approximate this objective through discrete metrics that can be efficiently computed. The performance improvement on recent instances $\mathcal{Q}_K$ is formulated as:

\begin{equation}
\small
    \Delta S(p',p) = \sum_{(x,y) \in \mathcal{Q}_K}A(p',p,x,y).
\label{eq:gain}
\end{equation}

\noindent The distribution ratio is approximated through correctness divergence:

\begin{equation}
\small
    D(p',p) = \sum_{(x,y) \in \mathcal{Q}_K} \mathbb{I}[f(p',x,y) \neq f(p,x,y)],
\label{eq:div}
\end{equation}

\noindent where $\mathbb{I}[\cdot]$ is the indicator function. We implement this objective by selecting candidates with positive $\Delta s$ and ranking them by performance improvement and divergence $(\Delta s, -d)$, as detailed in \Cref{alg:filter}. This lexicographic ranking considers distribution stability, ensuring prompt updates maintain consistency with existing knowledge while incorporating targeted improvements. The approach provides two key benefits: reduced computational cost compared to full validation set evaluation, and conservative prompt updates that limit the risk of degradation on unseen instances.

\subsection{Adaptive Knowledge Pruning}

\begin{algorithm}[t]
\caption{Violation Detection}
\label{alg:detection}
\small
\SetEndCharOfAlgoLine{}
\SetKwInOut{Input}{Input}
\SetKwInOut{Output}{Output}
\Input{
    Prompt $p$,
    Max children $C$,
    Max factor $F$
}
\Output{$issues_{local}$: Nodes with local degree violations, $issues_{global}$: Nodes with global balance violations}
\BlankLine
Parse prompt: $\mathcal{T} = (V,E) \leftarrow \text{Parse}(p)$ \\
Initialize $issues_{local} \leftarrow \emptyset, issues_{global} \leftarrow \emptyset$ \\
\For{$v \in V$ in pre-order traversal}{
    $\mathcal{T}_v \leftarrow$ subtree rooted at $v$ \\
    Calculate $\text{bf}(\mathcal{T}_v)$ using \Cref{eq:bf} \\
    \If{$\text{outdeg}(v) > C$}{
        $issues_{local} \leftarrow issues_{local} \cup \{v\}$
    }
    \If{$v \in V_t \land \beta(v) > F$}{
        $issues_{global} \leftarrow issues_{global} \cup \{v\}$
    }
}
\Return $(issues_{local}, issues_{global})$
\end{algorithm}

While knowledge integration improves performance, the iterative knowledge provision can lead to excessively long prompts that exceed practical token limits. We introduce a pruning mechanism by detecting structural inefficiencies in the hierarchy and providing guidance to $\text{LLM}_{opt}$ for targeted knowledge reduction. We define two types of structural constraints to identify pruning opportunities:

\textbf{Local Degree Constraint}: Limits the immediate breadth of any single category, effectively controlling the information density within a topic:
\begin{equation}
\small
    \forall v \in V_t: \text{outdeg}(v) \leq C,
\end{equation}
where $\text{outdeg}(v)$ accounts for both topic and note child nodes. This constraint forces the optimizer to consolidate or prune nodes to restore local manageability.

\textbf{Global Balance Constraint}: Ensures hierarchical efficiency by detecting structural anomalies where a topic node is too broad relative to its sub-topics. The constraint is defined as:
\begin{equation}
\small
    \forall v \in V_t: \beta(v) = \frac{\text{outdeg}(v)}{\text{bf}(\mathcal{T}_v)} \leq F,
\end{equation}

\noindent where the branching factor of subtree $\mathcal{T}_v$ is defined as:
\begin{equation}
\small
    \text{bf}(\mathcal{T}_v) = \frac{\sum_{u \in V_t(\mathcal{T}_v)} \text{outdeg}(u)}{|V_t(\mathcal{T}_v)|},
\label{eq:bf}
\end{equation}

\noindent and $F$ controls the maximum allowed deviation. A violation indicates that a parent topic is excessively flat compared to the depth of its children, requiring hierarchical reorganization. The detected nodes with violations provide explicit instructions to $\text{LLM}_{opt}$ in candidate prompt generation for performing targeted pruning. The violation detection procedure is detailed in \Cref{alg:detection}. This constraint-guided approach ensures that pruning maintains essential domain information while resolving organizational inefficiencies. 

\section{Experiments}

\subsection{Datasets}

To comprehensively evaluate KPPO's effectiveness on knowledge-intensive tasks, we curate a diverse benchmark suite of 15 tasks spanning multiple specialized domains where domain expertise and terminology precision are critical. Our selection criteria prioritize tasks that require substantial domain-specific knowledge beyond general language understanding. The selected benchmarks are organized into three domain categories:

\textbf{Financial Domain.} We employ FiQA \cite{fiqa}, a challenging sentiment analysis task requiring models to identify specific financial entities and classify sentiments toward multiple hierarchical aspects.

\textbf{Legal Domain.} From the LexGLUE benchmark \cite{lexglue}, we select the Case Hold task evaluating models' ability to identify correct legal holdings from court decisions. Each instance presents an excerpt from a court decision with a masked holding statement.

\textbf{Medical Domain.} We incorporate three medical benchmarks that collectively assess different facets of medical knowledge: MedQA \cite{medqa} (USMLE-style clinical vignettes), MedConceptsQA \cite{medconceptsqa} (medical coding and terminology), and MedMCQA \cite{medmcqa} (medical entrance exams). To assess fine-grained knowledge provision, we select 11 distinct tasks from MedMCQA: Anatomy, Dental, Gynecology \& Obstetrics, Medicine, Microbiology, Pathology, Pediatrics, Pharmacology, Physiology, Social \& Preventive Medicine, and Surgery.

\subsection{Baselines}
We evaluate our proposed framework against several well-established prompt optimization methods:

\textbf{Base Prompt}: We establish a minimal baseline using the system prompt ``You are a helpful assistant'' without any domain-specific content or task-specific instructions.

\textbf{Base Prompt (w/ cheat sheet)}: We employ the optimizer LLM to generate a concise summary of key concepts and rules derived from the first 10 training instances. This summary is prepended to the system prompt as a reference.

\textbf{RAG (w/ training set)}: we construct a local vector index of the entire training dataset. For each test query, we retrieve the top-k most similar training instances. These retrieved examples are prepended to the prompt as few-shot demonstrations.

\textbf{RAG (w/ web search)}: We implement a baseline where the task model is augmented with a search engine. The system performs a live web search, retrieves relevant documents, and appends the top-k snippets to the context window.

\textbf{Verify-and-Edit \cite{zhao2023verifyedit} (w/ web search)}: We implement a multi-stage pipeline where the model first generates a preliminary rationale, then generates search queries to verify and finally edits its response based on the retrieved evidence.

\textbf{OPRO} \cite{yang2024opro}: An algorithm that maintains a trajectory of historical prompts with their corresponding validation scores, using this meta-information to guide the generation of improved prompt candidates to maximize validation performance.

\textbf{APO} \cite{pryzant2023apo}: An approach that analyzes errors to generate ``gradients'' describing prompt deficiencies, guiding iterative prompt refinement. We evaluate two variants: the original implementation and an iteration-aligned variant matching our optimization budget for fair comparison.

\textbf{AMPO} \cite{yang2024ampo}: A multi-branched prompt optimization method that iteratively develops prompts with multiple branches to handle diverse patterns in complex tasks. This multi-branched design is particularly relevant for knowledge-intensive domains where tasks exhibit heterogeneous question types requiring different knowledge components.

\textbf{PromptAgent} \cite{wang2024promptagent}: A planning-based framework that formulates prompt optimization as sequential decision-making, employing Monte Carlo Tree Search (MCTS) to systematically explore the prompt space. We evaluate two variants: the original implementation and an iteration-aligned variant matching our optimization budget for fair comparison.

\subsection{Implementation Details}
We employ Llama~3.1-8B~\cite{dubey2024llama3.1} and Qwen~2.5-7B~\cite{qwen2025qwen2.5} as target language models with a sampling temperature of $0.8$, using vLLM with a specified seed for high-throughput inference. For the optimizer, we employ DeepSeek-V3 with a temperature of $1.0$ to encourage diverse candidate generation. To automatically map the prompt into the graph structure, we enforce a strict Markdown syntax in the optimizer's output, which is later parsed by the abstract syntax tree (AST). Training instances are selected via sentence embedding similarity~\cite{reimers2019sbert} to evaluation samples, ensuring semantic relevance for knowledge provision. Each task employs approximately 150 training samples, 50 validation samples, and 100 test samples; datasets lacking standard splits use validation sets as test sets. We utilize a batch size $B=5$, a recent instance window $K=10$, and a training budget of $T=60$ iterations. For the pruning variant, we set the maximum children per topic $C=8$ and the maximum balance factor $F=4.0$. Iteration-aligned baselines (APO*, PromptAgent*) are restricted to the same 60-iteration budget to ensure fair comparison.

\begin{table*}[t]
    \centering
    \setlength{\tabcolsep}{2pt}
    \renewcommand{\arraystretch}{1.05}
    \caption{Performance comparison (accuracy \%) of prompt optimization methods across 15 tasks from various domains.}
    \resizebox{\linewidth}{!}{
    \begin{threeparttable}
    \begin{tabular}{lcccccccccccccccc|c}
        \toprule
        \textbf{Method} && \textbf{FiQA} & \textbf{Case} & \textbf{Med} & \textbf{MedC} & \textbf{Anat} & \textbf{Dent} & \textbf{G\&O} & \textbf{Medn} & \textbf{Mic} & \textbf{Path} & \textbf{Ped} & \textbf{Phar} & \textbf{Phys} & \textbf{S\&P} & \textbf{Surg} & \textbf{AVG} \\
        \midrule
        \multicolumn{17}{c}{\textbf{Results on Llama 3.1}} \\
        \midrule
        Base && 73.2 & 49.0 & 69.0 & 28.3 & 61.3 & 47.0 & 65.1 & 63.4 & 61.0 & 71.3 & 62.1 & 76.9 & 66.7 & 53.9 & 53.0 & 60.1 \\
        Base (w/ Cheat Sheet) && \textbf{78.4} & 47.0 & 61.0 & 21.7 & 67.7 & 49.7 & 57.4 & 61.8 & 53.2 & 74.7 & 53.8 & 73.8 & 64.6 & 60.3 & 51.5 & 58.4 \\
        RAG (w/ Training Set) && 66.0 & 32.0 & 57.0 & 46.7 & 63.7 & 47.5 & 64.3 & 65.6 & 67.5 & 65.5 & 59.8 & 70.8 & 60.4 & 52.6 & 58.3 & 58.5 \\
        RAG (w/ Web Search) && 72.2 & 50.0 & 69.0 & 46.7 & 62.9 & 48.1 & 57.4 & 62.6 & 55.8 & 73.6 & 59.8 & 77.7 & 65.6 & 67.9 & 59.8 & 61.9 \\
        VE \cite{zhao2023verifyedit} (w/ Web Search) && 66.0 & \textbf{60.0} & 65.0 & 43.3 & \textbf{75.8} & 50.8 & 66.7 & 62.6 & 64.9 & 67.8 & 63.6& 70.0 & 65.6 & 62.8 & 54.5 & 62.6 \\
        OPRO \cite{yang2024opro} && 73.2 & 55.0 & \textbf{70.0} & 30.0 & 63.7 & 50.3 & 58.9 & 64.9 & 53.3 & 72.4 & \textbf{69.7} & 76.9 & 62.5 & 64.1 & 57.6 & 61.5 \\
        APO \cite{pryzant2023apo} && 73.2 & 52.0 & 65.0 & 35.0 & 61.3 & 45.9 & 61.2 & 63.4 & 59.7 & 71.3 & 65.2 & 74.6 & 66.7 & 61.5 & 60.6 & 61.1 \\
        APO* \cite{pryzant2023apo} && 73.2 & 55.0 & 63.0 & 33.3 & 66.1 & 51.4 & 57.4 & 63.4 & 58.4 & 71.3 & 57.6 & 73.1 & 62.5 & 53.9 & 56.8 & 59.8 \\
        AMPO \cite{yang2024ampo} && 73.2 & 49.0 & 65.0 & 28.3 & 60.5 & 45.3 & 62.8 & 63.4 & 61.0 & 72.4 & 66.7 & 76.2 & 60.4 & 53.9 & 62.1 & 60.0 \\
        PromptAgent \cite{wang2024promptagent} && 73.2 & 51.0 & 65.0 & 28.3 & 55.7 & 39.8 & 62.8 & 64.1 & \textbf{67.5} & 74.7 & 65.2 & 76.2 & 69.8 & 64.1 & 54.6 & 60.8 \\
        PromptAgent* \cite{wang2024promptagent} && 73.2 & 45.0 & 65.0 & 28.3 & 65.3 & 45.3 & 54.3 & 61.8 & 59.7 & 71.3 & 64.4 & 74.6 & 61.5 & 60.3 & 59.9 & 59.3 \\
        \textbf{KPPO (Ours)} && 77.3 & 55.0 & \textbf{70.0} & \textbf{51.7} & 71.0 & \textbf{58.6} & \textbf{69.0} & \textbf{68.7} & 61.0 & \textbf{75.9} & \textbf{69.7} & \textbf{80.8} & \textbf{71.9} & \textbf{69.2} & \textbf{63.6} & \textbf{67.6} \\
        \midrule
        \multicolumn{17}{c}{\textbf{Results on Qwen 2.5}} \\
        \midrule
        Base && 74.2 & 61.0 & 60.0 & 35.0 & 54.0 & 44.8 & 53.5 & 56.5 & 57.1 & 69.0 & 56.8 & 68.5 & 61.5 & \textbf{62.8} & 47.7 & 57.5 \\
        Base (w/ Cheat Sheet) && 76.3 & 57.0 & 62.0 & 30.0 & 58.0 & 43.6 & 51.2 & 53.4 & 54.5 & 69.0 & 59.8 & 66.9 & 63.5 & 61.5 & 49.2 & 57.0 \\
        RAG (w/ Training Set) && 74.2 & 47.0 & 57.0 & 41.7 & 50.0 & 45.9 & 45.0 & 48.1 & 65.0 & 52.9 & 51.5 & 64.6 & 60.4 & 53.9 & 38.6 & 53.0 \\
        RAG (w/ Web Search) && 74.2 & 61.0 & 64.0 & \textbf{53.3} & \textbf{66.1} & 47.0 & 50.4 & 61.1 & 58.4 & 70.1 & \textbf{60.6} & 72.3 & 61.5 & 56.4 & 43.2 & 60.0 \\
        VE \cite{zhao2023verifyedit} (w/ Web Search) && 76.3 & 61.0 & \textbf{65.0} & 43.3 & 62.1 & 52.5 & 52.7 & 56.5 & 57.1 & 72.4 & 52.3 & 76.2 & 63.5 & 61.5 & 48.5 & 60.1 \\
        OPRO \cite{yang2024opro} && 73.2 & 54.0 & 59.0 & 35.0 & 52.4 & 47.0 & 54.3 & 58.0 & \textbf{66.2} & 74.7 & 57.8 & 72.3 & \textbf{67.7} & 52.6 & 51.5 & 58.4 \\
        APO \cite{pryzant2023apo} && 75.3 & \textbf{62.0} & 55.0 & 30.0 & 54.8 & 45.9 & 50.4 & 53.4 & 57.1 & 64.4 & 53.0 & 71.5 & 63.5 & 55.1 & 47.7 & 55.9 \\
        APO* \cite{pryzant2023apo} && 76.3 & 57.0 & 64.0 & 35.0 & 55.7 & 45.3 & 58.1 & 53.4 & 53.3 & 63.2 & 53.8 & 66.2 & 61.5 & 59.0 & 46.2 & 56.5 \\
        AMPO \cite{yang2024ampo} && 75.3 & 61.0 & 64.0 & 20.0 & 58.1 & 44.8 & 53.5 & 54.2 & 57.1 & 65.5 & 53.0 & 69.2 & 65.6 & 55.1 & 52.3 & 56.6 \\
        PromptAgent \cite{wang2024promptagent} && 76.3 & 60.0 & 61.0 & 31.7 & 53.2 & 40.9 & 56.6 & 61.1 & 57.1 & 77.0 & 52.3 & 73.1 & 64.6 & 61.5 & 49.2 & 58.4 \\
        PromptAgent* \cite{wang2024promptagent} && 76.3 & 60.0 & 58.0 & 50.0 & 52.4 & 48.6 & 51.2 & 58.8 & 64.9 & 66.7 & 55.3 & 68.5 & 61.5 & 52.6 & \textbf{58.3} & 58.9 \\
        \textbf{KPPO (Ours)} && \textbf{77.3} & 61.0 & \textbf{65.0} & \textbf{53.3} & 60.5 & \textbf{56.4} & \textbf{58.9} & \textbf{66.4} & \textbf{66.2} & \textbf{80.5} & 59.8 & \textbf{80.8} & 66.7 & \textbf{62.8} & 57.6 & \textbf{64.9} \\
        \bottomrule
    \end{tabular}
        \begin{tablenotes}
            \scriptsize
            \item[]  Abbreviations: Case = Case Hold, Med = MedQA, MedC = MedConceptsQA, Anat = Anatomy, Dent = Dental, G\&O = Gynaecology \& Obstetrics, Medn = Medicine, Mic = Microbiology, Path = Pathology, Ped = Pediatrics, Phar = Pharmacology, Phys = Physiology, S\&P = Social \& Preventive Medicine, Surg = Surgery. `*' denotes methods with optimization iterations aligned to our framework.
       \end{tablenotes}
    \end{threeparttable}}
    \label{tab:main}
\end{table*}

\subsection{Quantitative Results}
\Cref{tab:main} presents comprehensive performance comparisons across all 15 knowledge-intensive benchmarks. KPPO achieves average accuracy of 67.6\% on Llama 3.1 and 64.9\% on Qwen 2.5, significantly surpassing not only traditional elicitation baselines where methods like OPRO achieve only marginal gains (+1.4\%) and others like APO frequently underperform the baseline, but also strong retrieval-augmented competitors, outperforming the tool-enhanced Verify-and-Edit (VE) method by +5.0\%. The superiority over retrieval-based methods validates that effective optimization requires not just accessing external knowledge, but synthesizing it into conflict-free logic. Traditional elicitation-based methods exhibit even smaller performance gain, confirming that elicitation-based methods struggle in knowledge-intensive tasks when clear task instructions are provided. The iteration-aligned variants yield suboptimal results and reveal architecture-dependent dynamics: extended iterations modestly improve baselines on Qwen 2.5 (APO*: +0.4\%, PromptAgent*: +1.4\%) but degrade performance on Llama 3.1 (APO*: -0.2\%, PromptAgent*: -1.5\%), suggesting overfitting to superficial patterns without genuine knowledge acquisition. These results validate our hypothesis that provision-based optimization transcends knowledge boundaries by directly augmenting models' effective knowledge bases, while elicitation-based approaches cannot overcome fundamental knowledge deficiencies.

\textbf{Discussion.} We specifically highlight performance on MedConceptsQA as a proxy for emerging domains where parametric knowledge is non-existent. While the base Llama 3.1 model achieves only 28.3\% accuracy due to these knowledge voids, KPPO improves performance to \textbf{51.7\%}. Notably, this outperforms the RAG baseline, demonstrating that in specialized domains where external documentation may be sparse or complex, KPPO's ability to synthesize domain rules directly from failure cases offers superior adaptability.

\subsection{Ablation Studies}

\begin{figure}[t]
  \centering
  \includegraphics[scale=0.25]{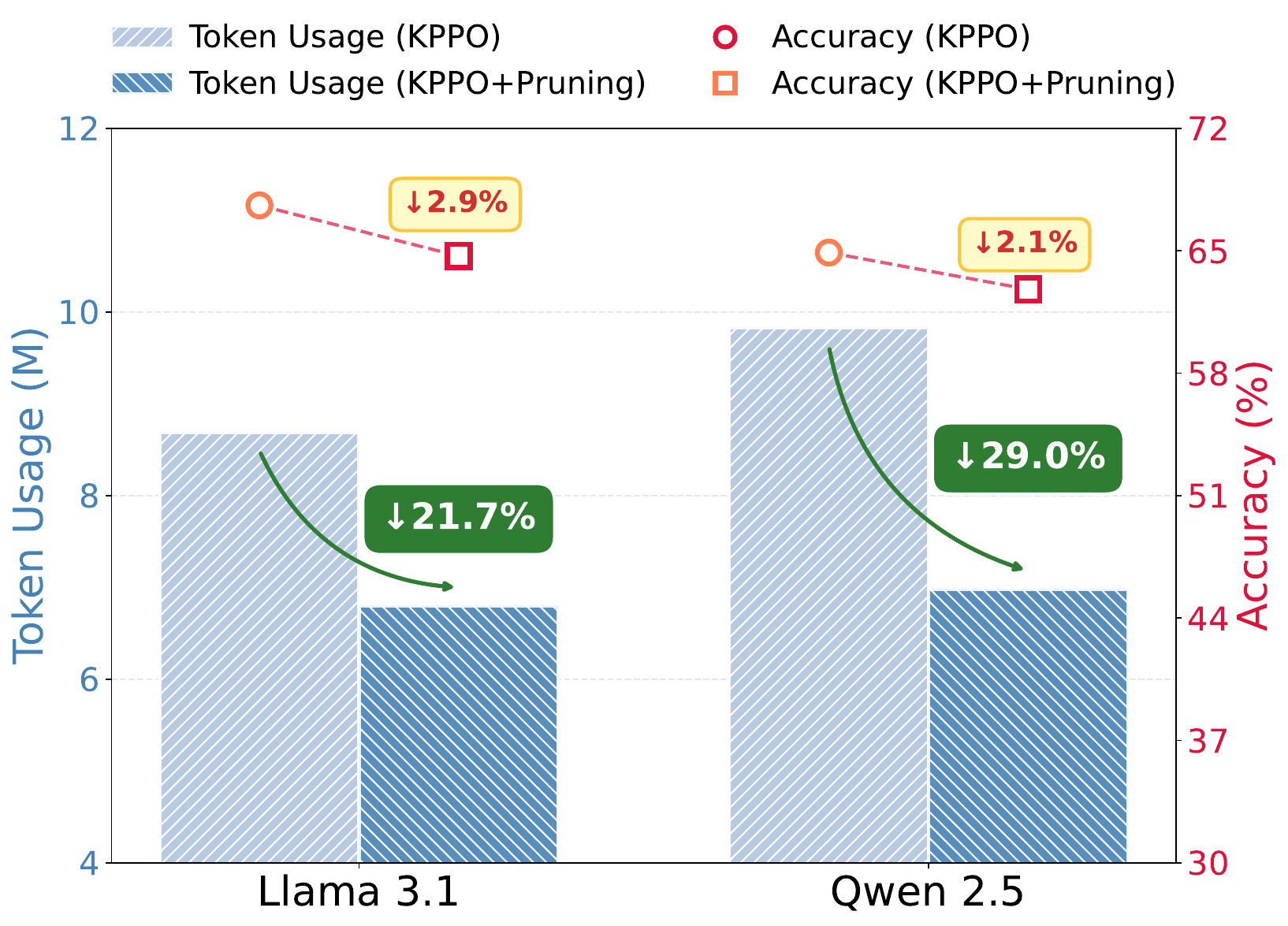}
  \caption{Inference token efficiency vs. performance trade-off of adaptive knowledge pruning on 15 tasks.}
  \label{fig:pruning}
\end{figure}

\noindent\textbf{Effect of Adaptive Knowledge Pruning.}
\Cref{fig:pruning} illustrates the efficiency-performance trade-off of adaptive pruning across all benchmarks. The pruning mechanism substantially reduces inference token consumption by 21.7\% for Llama~3.1 and 29.0\% for Qwen~2.5, decreasing average usage from 9.26M to 6.89M tokens. The corresponding accuracy drops are modest: 2.9\% for Llama~3.1 and 2.1\% for Qwen~2.5, demonstrating that our constraint-guided mechanism effectively identifies and removes organizational redundancy while preserving essential domain knowledge. Notably, Qwen~2.5 exhibits greater token reduction with smaller performance degradation, revealing architecture-dependent sensitivity to prompt structure and knowledge organization. Critically, the pruned variant maintains substantial superiority over all baseline methods, with margins of 3.2\% over the strongest competitor for Llama~3.1 and 3.9\% for Qwen~2.5. This validates our pruning strategy's ability to balance knowledge provision with inference token efficiency for target LLM, a critical consideration for practical deployment where token costs directly impact expenses.

\begin{figure*}[t]
  \centering
  \includegraphics[scale=0.35]{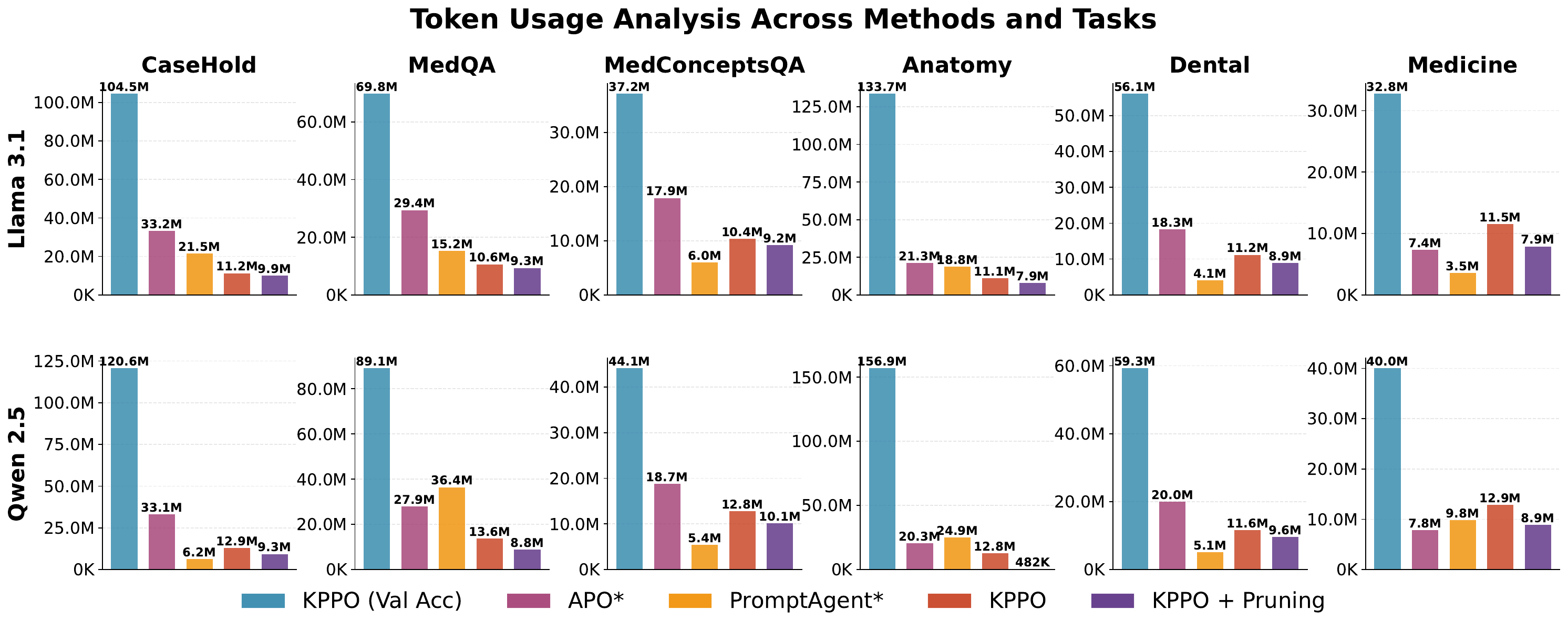}
  \caption{Token consumption analysis across optimization methods and KPPO variants. The `Val Acc' variant represents full validation set evaluation. `*' denotes baselines with optimization iterations aligned to our framework.}
  \label{fig:token}
\end{figure*}

\noindent\textbf{Token Efficiency Analysis.}
\Cref{fig:token} compares inference token utilization for target LLMs. Full validation evaluation that evaluates all candidates on the complete validation set (Val Acc) consumes substantially more tokens than elicitation-based methods due to provision-based prompts creating longer inputs and conducting comprehensive validation, with maximum disparity reaching 157M versus 20M tokens on Anatomy with Qwen2.5. KPPO with batch-wise validation demonstrates significantly lower consumption compared to full evaluation, establishing batch-wise evaluation as both effective and efficient. Compared to baselines, batch-wise KPPO exhibits competitive efficiency with PromptAgent and better efficiency than APO, despite the substantial improvement. The adaptive pruning mechanism further reduces token consumption across all tasks, achieving approximately 2M additional savings per task beyond batch-wise evaluation, demonstrating that our constraint-guided approach successfully balances knowledge provision with efficiency, a requirement for practical deployment where token costs scale with prompt length.

\begin{table}[t]
    \caption{Test-Time Computational Cost on Llama 3.1.}
    \centering
    \setlength{\tabcolsep}{1pt}
    \renewcommand{\arraystretch}{1.05}
    \resizebox{\linewidth}{!}{
    \begin{tabular}{lccc}
        \toprule
        \textbf{Method} & \textbf{Promp. Tokens} & \textbf{Comp. Tokens} & \textbf{Latency (s)} \\
        \midrule
        \midrule
        RAG (w/ Training Set) & \textbf{701} & 404 & 4.73 \\
        RAG (w/ Web Search) & 1172 & 385 & 36.16 \\
        VE (w/ Web Search) & 2803 & 770 & 37.98 \\
        KPPO (Ours) & 2782 & \textbf{377} & \textbf{3.88} \\
        \bottomrule
    \end{tabular}}
    \label{tab:inference}
\end{table}

\begin{table}[t]
    \caption{Ablation on candidate evaluation.}
    \centering
    \setlength{\tabcolsep}{4pt}
    \resizebox{\linewidth}{!}{
    \begin{threeparttable}
    \begin{tabular}{lccccccc}
        \toprule
        \textbf{Filtering} & \textbf{Case} & \textbf{Med} & \textbf{MedC} & \textbf{Anat} & \textbf{Dent} & \textbf{Medn} \\
        \midrule
        \multicolumn{7}{c}{\textbf{Results on Llama 3.1}} \\
        \midrule
        Val Acc & 50.0 & \textbf{70.0} & 36.7 & \textbf{71.8} & 49.2 & 59.5 \\
        Batch Acc & \underline{53.0} & 69.0 & \underline{40.0} & 66.1 & \underline{54.7} & \underline{61.8} \\
        Batch Acc + Div & \textbf{55.0} & \textbf{70.0} & \textbf{51.7} &\underline{71.0} & \textbf{58.6} & \textbf{68.7} \\
        \midrule
        \multicolumn{7}{c}{\textbf{Results on Qwen 2.5}} \\
        \midrule
        Val Acc & \textbf{63.0} & \underline{59.0} & 45.0 & 54.0 & \underline{49.2} & 59.5 \\
        Batch Acc & 54.0 & 56.0 & \underline{51.7} & \underline{58.9} & 48.6 & \underline{61.8} \\
        Batch Acc + Div & \underline{61.0} & \textbf{65.0} & \textbf{53.3} & \textbf{60.5} & \textbf{56.4} & \textbf{66.2} \\
        \bottomrule
    \end{tabular}
        \begin{tablenotes}
            \scriptsize
            \item[]  Abbreviations: Case = Case Hold, Med=MedQA, MedC = MedConceptsQA, Anat = Anatomy, Dent = Dental. Medn=Medicine.
       \end{tablenotes}
    \end{threeparttable}}
    \label{tab:ablation}
\end{table}

\noindent\textbf{Test-Time Inference Cost.}
To assess deployability, we quantified the averaged latency and token consumption of KPPO against retrieval-augmented baselines in \Cref{tab:inference}. As shown, KPPO demonstrates a decisive efficiency advantage, achieving an average latency of 3.88s per query on Llama 3.1. In contrast, web-search-augmented methods suffer from severe latency bottlenecks due to network overhead and multi-stage processing: RAG (w/ Web Search) requires 36.16s (9.3× slower), while Verify-and-Edit extends this to 37.98s (9.8× slower). Furthermore, KPPO significantly reduces costs, requiring only 377 completion tokens on average compared to 770 for Verify-and-Edit. This efficiency stems from KPPO's fundamental design: by bootstrapping complex reasoning and domain knowledge into a static prompt offline, the model can generate direct, accurate answers during inference.

\noindent\textbf{Candidate Evaluation Analysis.}
\Cref{tab:ablation} evaluates our dual-objective batch-wise evaluation mechanism against two alternatives: simple accuracy-based batch evaluation without divergence consideration (Batch Acc) and full validation evaluation (Val Acc). Our dual-objective approach (Batch Acc + Div) consistently outperforms simple batch-wise evaluation, achieving average improvements of 5.1\% on Llama~3.1 and 5.2\% on Qwen~2.5, demonstrating that maintaining distributional stability during prompt updates is critical for robust knowledge integration. Notably, full validation evaluation, despite its computational expense, often underperforms batch-wise methods. On MedConceptsQA, it falls 15.0\% and 8.3\% below dual-objective batch-wise evaluation for Llama~3.1 and Qwen~2.5 respectively. This substantial gap reveals that full validation optimization can encourage superficial pattern matching that improves global validation metrics without genuinely addressing knowledge gaps, whereas batch-wise evaluation better validates knowledge integration by focusing on resolving recent failure cases. The dual-objective criterion effectively prevents overfitting to spurious patterns by penalizing excessive distributional changes, ensuring that selected prompts provide generalizable domain knowledge rather than instance-specific adjustments. These results validate our design choice to prioritize targeted knowledge gap remediation over global validation performance maximization.

\begin{figure*}[t]
  \centering
  \includegraphics[scale=0.38]{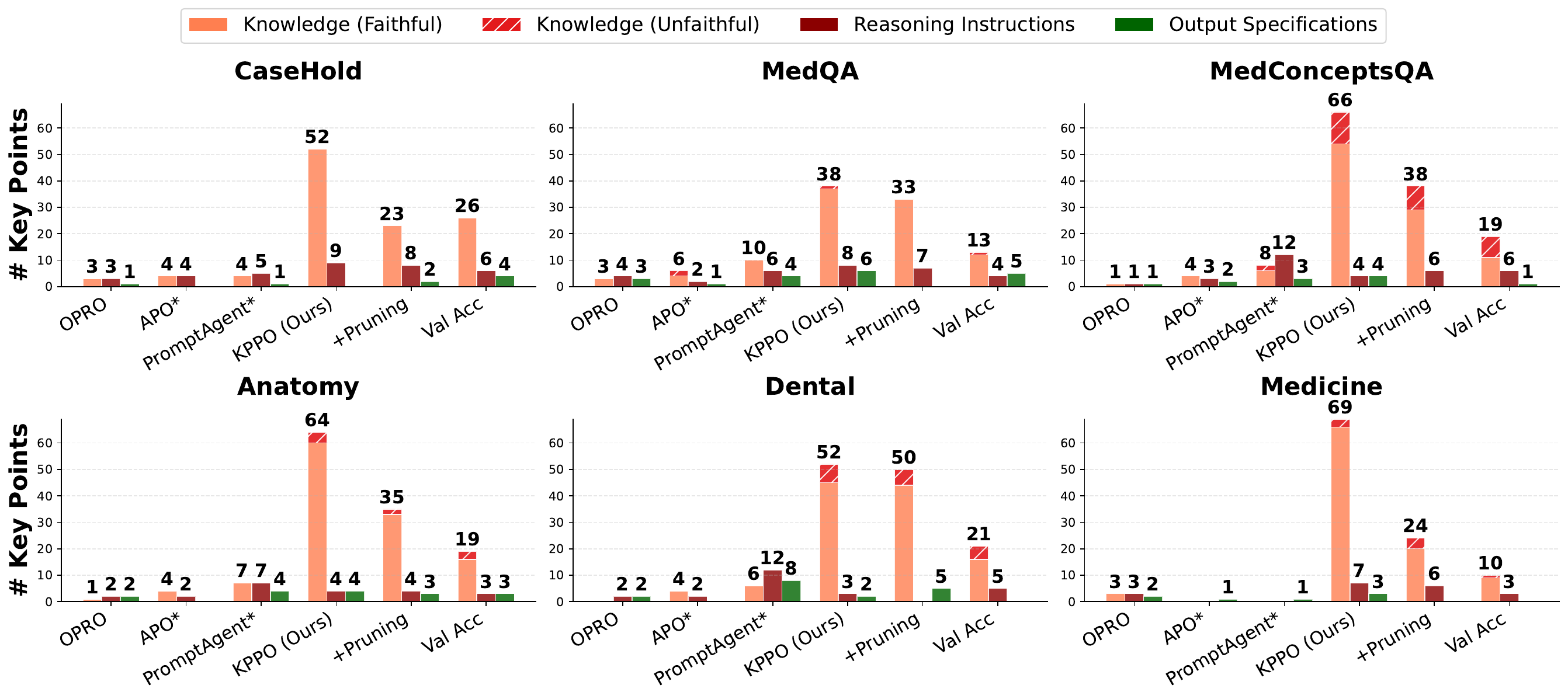}
  \caption{Quantitative analysis of knowledge content in optimized prompts across six tasks on Llama~3.1, categorized into domain knowledge, reasoning instructions, and output specifications.}
  \label{fig:keypoint}
\end{figure*}

\begin{figure}[t]
  \centering
  \includegraphics[scale=0.3]{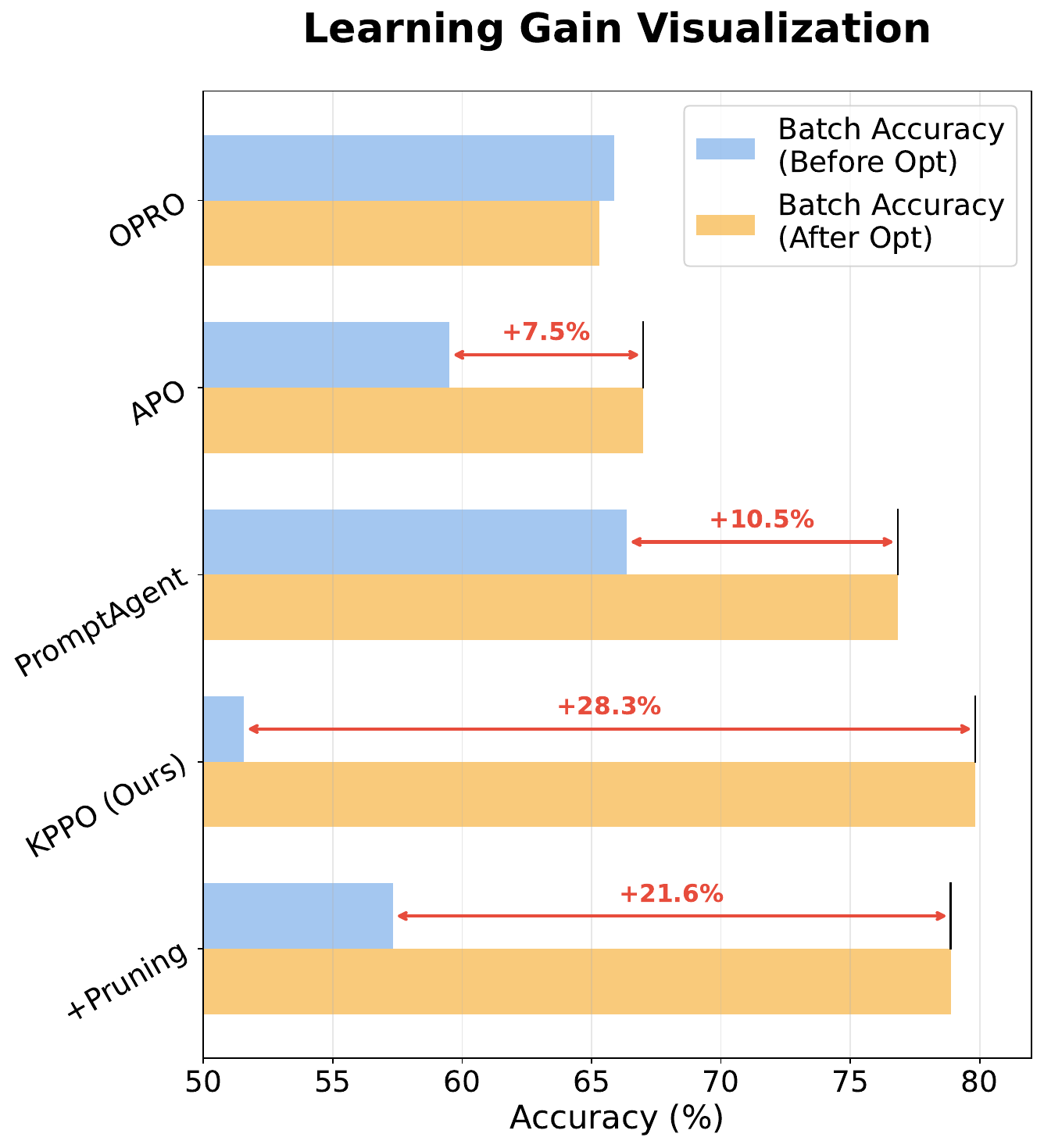}
  \caption{Learning gain comparison across optimization methods with Llama.}
  \label{fig:learning_gain}
\end{figure}

\noindent\textbf{Prompt Knowledge Analysis.}
\Cref{fig:keypoint} presents quantitative analysis of knowledge content across three categories: domain knowledge, reasoning instructions, and output specifications. To assess knowledge quality, we conducted a faithfulness audit using a superior online LLM (Gemini 3 Pro) coupled with human verification. KPPO integrates substantially more domain knowledge than elicitation-based methods, averaging 56.8 points on Llama~3.1 (26.2$\times$ more than OPRO and 7.4$\times$ more than PromptAgent), directly addressing the knowledge poverty limitation of elicitation approaches. Notably, KPPO maintains high fidelity, with an average faithfulness rate of 92.5\% across evaluated tasks. The adaptive pruning mechanism effectively reduces knowledge points by 30-50\% across tasks while maintaining essential information. Critically, full validation evaluation consistently yields lower domain knowledge content than even the pruned variant, with particularly differences on MedConceptsQA (19 vs. 38 points), revealing that optimizing for global validation metrics does not encourage substantive knowledge integration and leads to lower faithfulness (83.4\%). This disparity reveals that optimizing solely for global validation metrics can encourage the generation of spurious correlations that fit the validation set but lack factual grounding. These findings validate our batch-wise evaluation design, which prioritizes failure case resolution over validation accuracy maximization, ensuring that optimization genuinely expands the model's effective knowledge base.

\noindent\textbf{Optimization Effectiveness Analysis.}
\Cref{fig:learning_gain} presents optimization effectiveness across methods through the learning gain metric, which measure the proportion of previous failure cases successfully resolved during optimization. KPPO achieves 28.3\% learning gain, substantially outperforming elicitation-based baselines: 3.8$\times$ higher than APO (7.5\%) and 2.7$\times$ higher than PromptAgent (10.5\%). The pruning variant maintains 21.6\% learning gain, still exceeding all baselines by substantial margins. These results demonstrate KPPO's superior capability in addressing knowledge gaps through targeted knowledge provision rather than superficial pattern adjustments. KPPO also exhibits lower initial batch accuracy compared to baselines, which we attribute to the introduction of provisional knowledge that may initially introduce noise; however, through rigorous iterative refinement and batch-wise validation, this provisional knowledge converges to reliable domain expertise that systematically resolves failure cases.

\begin{figure*}[t]
  \centering
  \begin{minipage}[t]{0.59\linewidth}
    \centering
    \includegraphics[width=\linewidth]{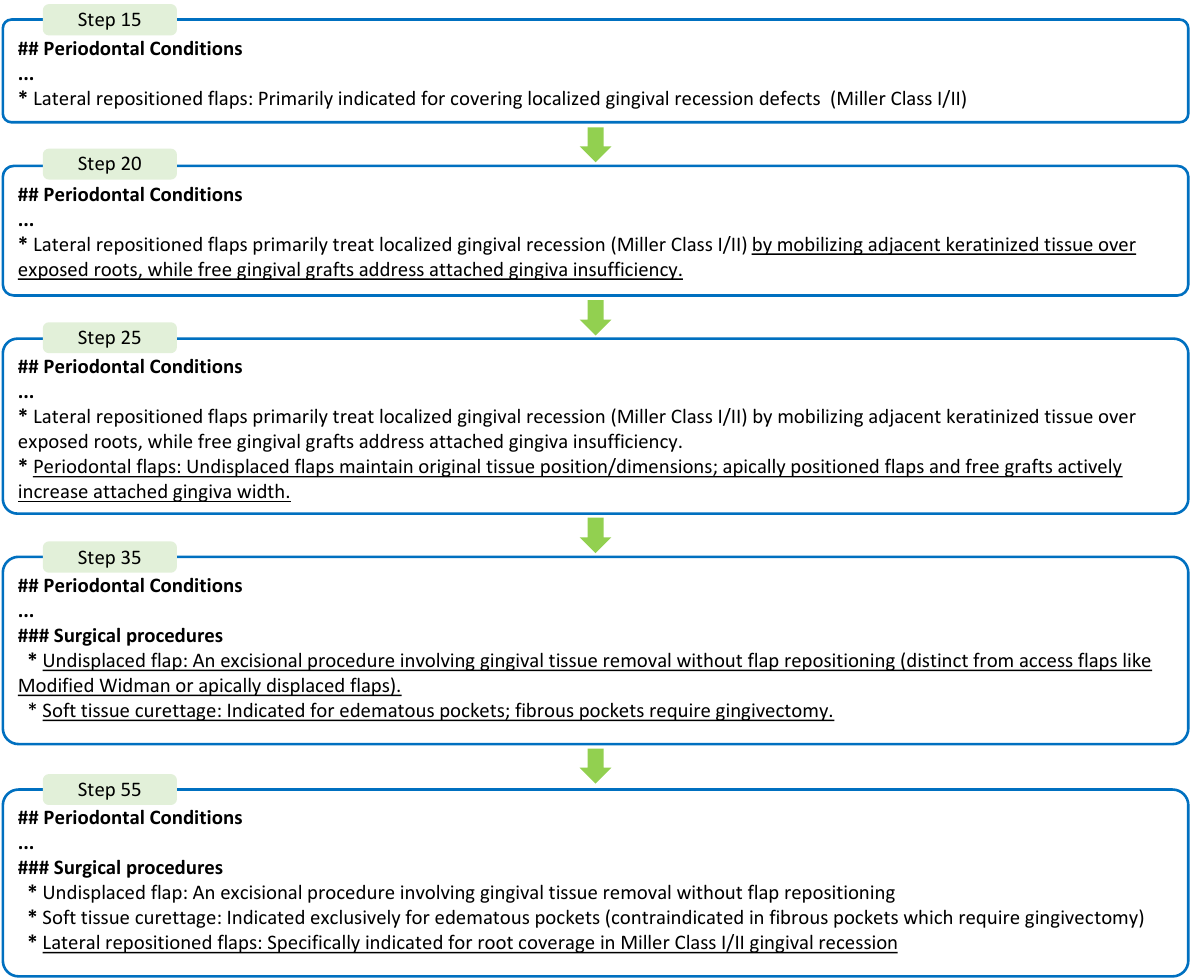}
    \caption{Progressive knowledge accumulation in periodontal surgical procedures across six optimization steps. Each panel shows incremental additions (marked with underline) to the knowledge base. This pattern exemplifies monotonic knowledge growth typical of well-structured domain ontologies.}
    \label{fig:provision}
  \end{minipage}
  \hfill
  \begin{minipage}[t]{0.38\linewidth}
    \centering
    \includegraphics[width=\linewidth]{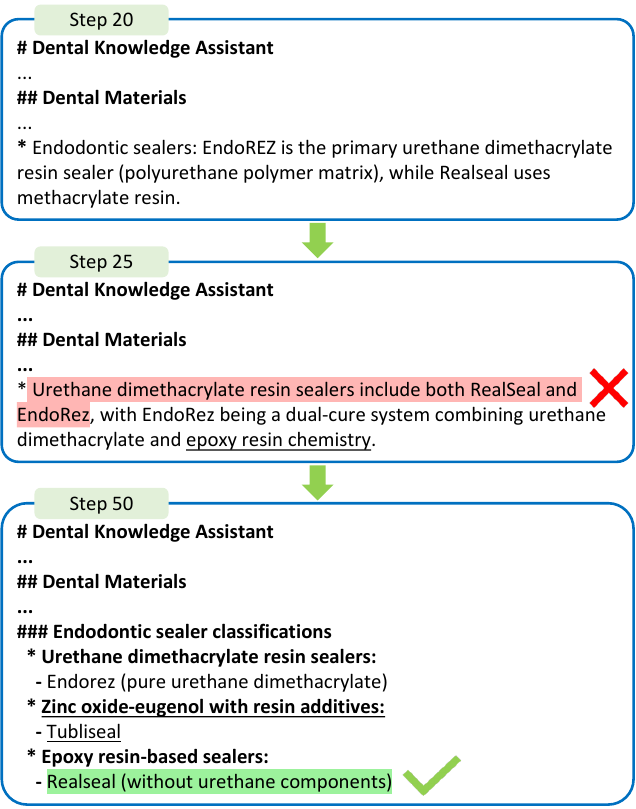}
    \caption{Self-correction trajectory demonstrating non-monotonic accuracy evolution. Step 25 introduces a taxonomic error (red). Step 50 implements correction through structured taxonomy and negative specifications (green).}
    \label{fig:correction}
  \end{minipage}
\end{figure*}

\begin{figure*}[t]
  \centering
  \includegraphics[scale=0.54]{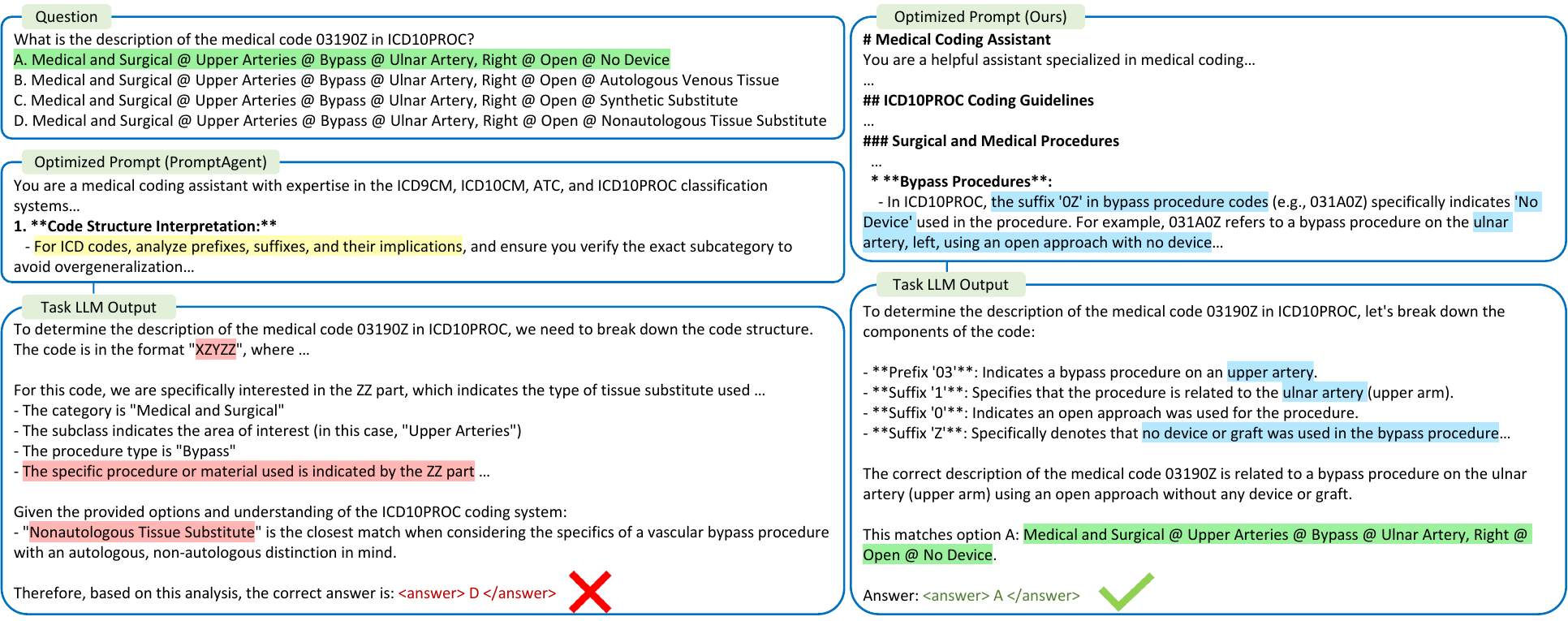}
  \caption{Qualitative comparison between PromptAgent and KPPO on medical code interpretation task. PromptAgent provides generic analytical guidance (yellow highlighting) that fails to address the specific knowledge gap, resulting in incorrect code interpretation (red answer). KPPO integrates domain knowledge about ICD10PROC conventions (blue highlighting), enabling accurate identification that suffix `0Z' indicates `No Device', leading to the correct answer (green).}
  \label{fig:case_study}
\end{figure*}

\subsection{Qualitative Results}

To understand how prompts evolve through iterative refinement, we trace the longitudinal development of specific knowledge components across optimization steps.

\noindent\textbf{Knowledge Provision.}
\Cref{fig:provision} illustrates monotonic knowledge expansion through the evolution of periodontal surgical procedures across six optimization checkpoints. The trajectory demonstrates three key characteristics of effective knowledge provision: hierarchical expansion from isolated procedural descriptions to comprehensive taxonomically-organized collections encompassing multiple surgical modalities; dimensional augmentation that introduces orthogonal clinical aspects such as procedural mechanisms (Step 20), flap type classifications (Step 25), and structured taxonomies with contraindications (Step 35-55); and specificity refinement evident in evolution from general statements (``indicated for edematous pockets'') to precise clinical specifications (``indicated exclusively for edematous pockets''). This accumulation pattern is particularly valuable for protocol-driven domains where systematic knowledge expansion occurs through iterative gap-filling based on failure case feedback, enabling comprehensive coverage without disrupting previously validated content.

\noindent\textbf{Self-Correction.}
In contrast to monotonic accumulation, \Cref{fig:correction} reveals KPPO's capacity for self-diagnosis and rectification of domain-specific misconceptions through endodontic sealer classification evolution. The trajectory exhibits non-monotonic accuracy dynamics: Step 20 correctly identifies EndoREZ as urethane dimethacrylate-based, Step 25 paradoxically introduces taxonomic error by incorrectly grouping RealSeal with urethane-based sealers despite RealSeal being epoxy resin-based, and Step 50 implements systematic correction through structured taxonomy with explicit negative specifications (``RealSeal: epoxy resin-based, without urethane components''). This self-correction pattern demonstrates two critical insights: iterative knowledge provision can temporarily degrade accuracy when integrating unfaithful content, as the optimizer explores knowledge space based on limited failure cases; however, through continued refinement across diverse training batches, contradictions surface and the framework converges to faithful representations that resolve both original and newly-encountered failure cases.

\noindent\textbf{Baseline Comparison.}
\Cref{fig:case_study} provides qualitative comparison between PromptAgent and KPPO on a medical code interpretation task from MedConceptsQA. The baseline prompt offers generic guidance (``analyze prefixes, suffixes, and their implications'') that fails to address specific knowledge gaps, resulting in incorrect interpretation of ICD10PROC code 03190Z. In contrast, KPPO integrates precise domain knowledge about ICD10PROC coding conventions, specifically that ``the suffix `0Z' indicates `No Device' used in the procedure.'' This case exemplifies how KPPO addresses deficiencies through knowledge provision rather than superficial instruction reformulation. The qualitative analysis demonstrates KPPO's fundamental advantage: rather than hoping that better instructions can elicit non-existent knowledge, our framework directly provides the domain-specific information required for accurate task performance.

\section{Conclusion}

This paper addresses a fundamental limitation in automated prompt optimization: the inability of elicitation-based methods to transcend knowledge boundaries in knowledge-intensive domains. We introduce KPPO, a principled framework that reformulates prompt optimization as systematic knowledge integration through three technical innovations: knowledge gap filling via gradient-based failure analysis, batch-wise dual-objective evaluation ensuring robust integration over superficial pattern matching, and adaptive pruning optimizing computational cost through structural constraint guidance. Extensive evaluation across 15 benchmarks in financial, legal, and medical domains demonstrates KPPO's substantial improvements of approximately 6\% over strongest elicitation-based baselines, with the pruning variant achieving 21.7-29.0\% token reduction. These results establish provision-based optimization as a viable paradigm for deploying LLMs in specialized domains where task success fundamentally depends on domain expertise beyond knowledge capacity, providing an efficient alternative when traditional fine-tuning is computationally prohibitive or data-limited.

\bibliographystyle{IEEEtran}
\bibliography{references}

@inproceedings{wei2022cot,
title={Chain-of-thought prompting elicits reasoning in large language models},
author={Wei, Jason and Wang, Xuezhi and Schuurmans, Dale and Bosma, Maarten and Xia, Fei and Chi, Ed and Le, Quoc V and Zhou, Denny and others},
booktitle={NeurIPS},
volume={35},
pages={24824--24837},
year={2022}
}

@inproceedings{
zhang2023autocot,
title={Automatic chain of thought prompting in large language models},
author={Zhuosheng Zhang and Aston Zhang and Mu Li and Alex Smola},
booktitle={ICLR},
year={2023}
}

@inproceedings{brown2020gpt3,
title={Language models are few-shot learners},
author={Brown, Tom and Mann, Benjamin and Ryder, Nick and Subbiah, Melanie and Kaplan, Jared D and Dhariwal, Prafulla and Neelakantan, Arvind and Shyam, Pranav and Sastry, Girish and Askell, Amanda and others},
booktitle={NeurIPS},
volume={33},
pages={1877--1901},
year={2020}
}

@inproceedings{zhao2021calibrate,
title={Calibrate before use: Improving few-shot performance of language models},
author={Zhao, Zihao and Wallace, Eric and Feng, Shi and Klein, Dan and Singh, Sameer},
booktitle={ICML},
pages={12697--12706},
year={2021},
organization={PMLR}
}

@inproceedings{lu2022fantastically,
title={Fantastically ordered prompts and where to find them: Overcoming few-shot prompt order sensitivity},
author={Lu, Yao and Bartolo, Max and Moore, Alastair and Riedel, Sebastian and Stenetorp, Pontus},
booktitle={ACL},
pages={8086--8098},
year={2022}
}

@inproceedings{
zhou2023leasttomost,
title={Least-to-most prompting enables complex reasoning in large language models},
author={Denny Zhou and Nathanael Sch{\"a}rli and Le Hou and Jason Wei and Nathan Scales and Xuezhi Wang and Dale Schuurmans and Claire Cui and Olivier Bousquet and Quoc V Le and Ed H. Chi},
booktitle={ICLR},
year={2023}
}

@inproceedings{
zhang2023tempera,
title={Tempera: Test-time prompt editing via reinforcement learning},
author={Tianjun Zhang and Xuezhi Wang and Denny Zhou and Dale Schuurmans and Joseph E. Gonzalez},
booktitle={ICLR},
year={2023}
}

@inproceedings{prasad2023grips,
title={Grips: Gradient-free, edit-based instruction search for prompting large language models},
author={Prasad, Archiki and Hase, Peter and Zhou, Xiang and Bansal, Mohit},
booktitle={EACL},
pages={3845--3864},
year={2023}
}

@inproceedings{sorensen2022informationtheoretic,
title={An information-theoretic approach to prompt engineering without ground truth labels},
author={Sorensen, Taylor and Robinson, Joshua and Rytting, Christopher and Shaw, Alexander and Rogers, Kyle and Delorey, Alexia and Khalil, Mahmoud and Fulda, Nancy and Wingate, David},
booktitle={ACL},
pages={819--862},
year={2022}
}

@inproceedings{cheng2023bpo,
title = {Black-box prompt optimization: Aligning large language models without model training},
author = {Cheng, Jiale and Liu, Xiao and Zheng, Kehan and Ke, Pei and Wang, Hongning and Dong, Yuxiao  and Tang, Jie and Huang, Minlie},
booktitle = {ACL},
year = {2024},
pages = {3201--3219}
}

@inproceedings{kwon2024stableprompt,
title={Stableprompt: Automatic prompt tuning using reinforcement learning for large language model},
author={Kwon, Minchan and Kim, Gaeun and Kim, Jongsuk and Lee, Haeil and Kim, Junmo},
booktitle={EMNLP},
pages={9868--9884},
year={2024}
}

@inproceedings{
sun2024querydependent,
title={Query-dependent prompt evaluation and optimization with offline inverse rl},
author={Hao Sun and Alihan H{\"u}y{\"u}k and Mihaela van der Schaar},
booktitle={ICLR},
year={2024}
}

@inproceedings{
zhou2022ape,
title={Large language models are human-level prompt engineers},
author={Yongchao Zhou and Andrei Ioan Muresanu and Ziwen Han and Keiran Paster and Silviu Pitis and Harris Chan and Jimmy Ba},
booktitle={ICLR},
year={2023},
}

@inproceedings{pryzant2023apo,
title={Automatic prompt optimization with “gradient descent” and beam search},
author={Pryzant, Reid and Iter, Dan and Li, Jerry and Lee, Yin and Zhu, Chenguang and Zeng, Michael},
booktitle={EMNLP},
pages={7957--7968},
year={2023}
}

@inproceedings{
chen2024instructzero,
title={Instructzero: Efficient instruction optimization for black-box large language models},
author={Lichang Chen and Jiuhai Chen and Tom Goldstein and Heng Huang and Tianyi Zhou},
booktitle={ICML},
year={2024}
}

@inproceedings{
yang2024opro,
title={Large language models as optimizers},
author={Chengrun Yang and Xuezhi Wang and Yifeng Lu and Hanxiao Liu and Quoc V Le and Denny Zhou and Xinyun Chen},
booktitle={ICLR},
year={2024}
}

@inproceedings{
guo2024evoprompt,
title={Connecting large language models with evolutionary algorithms yields powerful prompt optimizers},
author={Qingyan Guo and Rui Wang and Junliang Guo and Bei Li and Kaitao Song and Xu Tan and Guoqing Liu and Jiang Bian and Yujiu Yang},
booktitle={ICLR},
year={2024}
}

@inproceedings{
fernando2024promptbreeder,
title={Promptbreeder: Self-referential self-improvement via prompt evolution},
author={Chrisantha Fernando and Dylan Sunil Banarse and Henryk Michalewski and Simon Osindero and Tim Rockt{\"a}schel},
booktitle={ICML},
year={2024}
}

@inproceedings{
lin2023instinct,
title={Use your instinct: Instruction optimization using neural bandits coupled with transformers},
author={Xiaoqiang Lin and Zhaoxuan Wu and Zhongxiang Dai and Wenyang Hu and Yao Shu and See-Kiong Ng and Patrick Jaillet and Bryan Kian Hsiang Low},
booktitle={NeurIPS Workshop on Instruction Tuning and Instruction Following},
year={2023}
}

@inproceedings{
wang2024promptagent,
title={Promptagent: Strategic planning with language models enables expert-level prompt optimization},
author={Xinyuan Wang and Chenxi Li and Zhen Wang and Fan Bai and Haotian Luo and Jiayou Zhang and Nebojsa Jojic and Eric Xing and Zhiting Hu},
booktitle={ICLR},
year={2024}
}

@inproceedings{ye2023pe2,
title = {Prompt engineering a prompt engineer},
author = {Ye, Qinyuan and Ahmed, Mohamed and Pryzant, Reid and Khani, Fereshte},
booktitle = {ACL Findings},
year = {2024},
pages = {355--385}
}

@inproceedings{long2024advicl,
title={Prompt optimization via adversarial in-context learning},
author={Long, Do and Zhao, Yiran and Brown, Hannah and Xie, Yuxi and Zhao, James and Chen, Nancy and Kawaguchi, Kenji and Shieh, Michael and He, Junxian},
booktitle={ACL},
pages={7308--7327},
year={2024}
}

@inproceedings{zhang2024glape,
title = "Glape: Gold label-agnostic prompt evaluation for large language models",
author = "Zhang, Xuanchang and Zhang, Zhuosheng and Zhao, Hai",
booktitle = "EMNLP",
year = "2024",
pages = "2027--2039",
}

@inproceedings{
shi2024triple,
title={Efficient prompt optimization through the lens of best arm identification},
author={Chengshuai Shi and Kun Yang and Zihan Chen and Jundong Li and Jing Yang and Cong Shen},
booktitle={NeurIPS},
year={2024}
}

@inproceedings{yang2024ampo,
title={Ampo: Automatic multi-branched prompt optimization},
author={Yang, Sheng and Wu, Yurong and Gao, Yan and Zhou, Zineng and Zhu, Bin and Sun, Xiaodi and Lou, Jian-Guang and Ding, Zhiming and Hu, Anbang and Fang, Yuan and others},
booktitle={EMNLP},
pages={20267--20279},
year={2024}
}

@article{agarwal2024promptwizard,
title={Promptwizard: Task-aware prompt optimization framework},
author={Agarwal, Eshaan and Singh, Joykirat and Dani, Vivek and Magazine, Raghav and Ganu, Tanuja and Nambi, Akshay},
journal={arXiv preprint arXiv:2405.18369},
year={2024}
}

@article{zhang2025mars,
title={Mars: A multi-agent framework incorporating socratic guidance for automated prompt optimization},
author={Zhang, Jian and Wang, Zhangqi and Zhu, Haiping and Liu, Jun and Lin, Qika and Cambria, Erik},
journal={arXiv preprint arXiv:2503.16874},
year={2025}
}

@inproceedings{choi2025system,
title={System prompt optimization with meta-learning},
author={Choi, Yumin and Baek, Jinheon and Hwang, Sung Ju},
booktitle={NeurIPS},
year={2025}
}

@article{hazman2025grammar,
title={Grammar-guided evolutionary search for discrete prompt optimisation},
author={Hazman, Muzhaffar and Pham, Minh-Khoi and Soundararajan, Shweta and Mordido, Goncalo and Custode, Leonardo and Lynch, David and Cruciata, Giorgio and Shi, Yucheng and Song, Hongmeng and Chao, Wang and others},
journal={arXiv preprint arXiv:2507.10326},
year={2025}
}

@article{zhao2025pmpo,
title={Pmpo: Probabilistic metric prompt optimization for small and large language models},
author={Zhao, Chenzhuo and Liu, Ziqian and Wang, Xingda and Lu, Junting and Ruan, Chaoyi},
journal={arXiv preprint arXiv:2505.16307},
year={2025}
}

@inproceedings{cui2025see,
title={See: Strategic exploration and exploitation for cohesive in-context prompt optimization},
author={Cui, Wendi and Zhang, Jiaxin and Li, Zhuohang and Sun, Hao and Lopez, Damien and Das, Kamalika and Malin, Bradley A and Kumar, Sricharan},
booktitle={ACL},
pages={29575--29627},
year={2025}
}

@inproceedings{kong2024prewrite,
title = "Prewrite: Prompt rewriting with reinforcement learning",
author = "Kong, Weize and Hombaiah, Spurthi and Zhang, Mingyang and Mei, Qiaozhu and Bendersky, Michael",
booktitle = "ACL",
year = "2024",
pages = "594--601"
}

@inproceedings{chen2024promst,
title={Prompt optimization in multi-step tasks (promst): Integrating human feedback and heuristic-based sampling},
author={Chen, Yongchao and Arkin, Jacob and Hao, Yilun and Zhang, Yang and Roy, Nicholas and Fan, Chuchu},
booktitle={EMNLP},
pages={3859--3920},
year={2024}
}

@inproceedings{zhang2024agentpro,
title = "Agent-pro: Learning to evolve via policy-level reflection and optimization",
author = "Zhang, Wenqi and Tang, Ke and Wu, Hai and Wang, Mengna and Shen, Yongliang and Hou, Guiyang  and Tan, Zeqi and Li, Peng and Zhuang, Yueting and Lu, Weiming",
booktitle = "ACL",
year = "2024",
pages = "5348--5375",
}

@inproceedings{ling2024deductive,
title={Deductive verification of chain-of-thought reasoning},
author={Ling, Zhan and Fang, Yunhao and Li, Xuanlin and Huang, Zhiao and Lee, Mingu and Memisevic, Roland and Su, Hao},
booktitle={NeurIPS},
volume={36},
year={2024}
}

@inproceedings{
gou2024critic,
title={Critic: Large language models can self-correct with tool-interactive critiquing},
author={Zhibin Gou and Zhihong Shao and Yeyun Gong and yelong shen and Yujiu Yang and Nan Duan and Weizhu Chen},
booktitle={ICLR},
year={2024}
}

@inproceedings{zhao2023verifyedit,
title = "Verify-and-edit: A knowledge-enhanced chain-of-thought framework",
author = "Zhao, Ruochen and Li, Xingxuan and Joty, Shafiq and Qin, Chengwei and Bing, Lidong",
booktitle = "ACL",
year = "2023",
pages = "5823--5840"
}

@inproceedings{madaan2024selfrefine,
title={Self-refine: Iterative refinement with self-feedback},
author={Madaan, Aman and Tandon, Niket and Gupta, Prakhar and Hallinan, Skyler and Gao, Luyu and Wiegreffe, Sarah and Alon, Uri and Dziri, Nouha and Prabhumoye, Shrimai and Yang, Yiming and others},
booktitle={NeurIPS},
volume={36},
year={2024}
}

@inproceedings{shinn2024reflexion,
title={Reflexion: Language agents with verbal reinforcement learning},
author={Shinn, Noah and Cassano, Federico and Gopinath, Ashwin and Narasimhan, Karthik and Yao, Shunyu},
booktitle={NeurIPS},
volume={36},
year={2024}
}

@article{pan2024survey,
title = "Automatically correcting large language models: Surveying the landscape of diverse automated correction strategies",
author = "Pan, Liangming and Saxon, Michael and Xu, Wenda and Nathani, Deepak and Wang, Xinyi and Wang, William Yang",
journal = "TACL",
volume = "12",
year = "2024",
pages = "484--506"
}

@inproceedings{
welleck2023selfcorrect,
title={Generating sequences by learning to self-correct},
author={Sean Welleck and Ximing Lu and Peter West and Faeze Brahman and Tianxiao Shen and Daniel Khashabi and Yejin Choi},
booktitle={ICLR},
year={2023}
}

@article{yuksekgonul2024textgrad,
title={Optimizing generative AI by backpropagating language model feedback},
author={Yuksekgonul, Mert and Bianchi, Federico and Boen, Joseph and Liu, Sheng and Lu, Pan and Huang, Zhi and Guestrin, Carlos and Zou, James},
journal={Nature},
volume={639},
number={8055},
pages={609--616},
year={2025},
publisher={Nature Publishing Group}
}

@inproceedings{trpo,
title={Trust region policy optimization},
author={Schulman, John and Levine, Sergey and Abbeel, Pieter and Jordan, Michael and Moritz, Philipp},
booktitle={ICML},
pages={1889--1897},
year={2015},
volume={37}
}

@inproceedings{fiqa,
title={WWW'18 open challenge: Financial opinion mining and question answering},
author={Maia, Macedo and Handschuh, Siegfried and Freitas, Andr{\'e} and Davis, Brian and McDermott, Ross and Zarrouk, Manel and Balahur, Alexandra},
booktitle={WWW},
pages={1941--1942},
year={2018}
}

@inproceedings{lexglue,
title={Lexglue: A benchmark dataset for legal language understanding in english},
author={Chalkidis, Ilias and Jana, Abhik and Hartung, Dirk and Bommarito, Michael and Androutsopoulos, Ion and Katz, Daniel and Aletras, Nikolaos},
booktitle={ACL},
pages={4310--4330},
year={2022}
}

@inproceedings{medmcqa,
title={Medmcqa: A large-scale multi-subject multi-choice dataset for medical domain question answering},
author={Pal, Ankit and Umapathi, Logesh Kumar and Sankarasubbu, Malaikannan},
booktitle={CHIL},
pages={248--260},
year={2022}
}

@article{medconceptsqa,
title={Medconceptsqa: Open source medical concepts qa benchmark},
author={Shoham, Ofir Ben and Rappoport, Nadav},
journal={CIBM},
volume={182},
pages={109089},
year={2024},
publisher={Elsevier}
}

@article{medqa,
title={What disease does this patient have? A large-scale open domain question answering dataset from medical exams},
author={Jin, Di and Pan, Eileen and Oufattole, Nassim and Weng, Wei-Hung and Fang, Hanyi and Szolovits, Peter},
journal={Appl. Sci.},
volume={11},
number={14},
pages={6421},
year={2021},
publisher={MDPI}
}

@inproceedings{reimers2019sbert,
title={Sentence-bert: Sentence embeddings using siamese bert-networks},
author={Reimers, Nils and Gurevych, Iryna},
booktitle={EMNLP},
pages={3982--3992},
year={2019}
}

@article{dubey2024llama3.1,
title={The llama 3 herd of models},
author={Dubey, Abhimanyu and Jauhri, Abhinav and Pandey, Abhinav and Kadian, Abhishek and Al-Dahle, Ahmad and Letman, Aiesha and Mathur, Akhil and Schelten, Alan and Yang, Amy and Fan, Angela and others},
journal={arXiv preprint arXiv:2407.21783},
year={2024}
}

@article{qwen2025qwen2.5,
title={Qwen2.5 technical report},
author={Yang, An and Li, Anfeng and Yang, Baosong and Zhang, Beichen and Hui, Binyuan and Zheng, Bo and Yu, Bowen and Gao, Chang and Huang, Chengen and Lv, Chenxu and others},
journal={arXiv preprint arXiv:2412.15115},
year={2025}
}

@article{liu2024deepseek,
title={Deepseek-v3 technical report},
author={Liu, Aixin and Feng, Bei and Xue, Bing and Wang, Bingxuan and Wu, Bochao and Lu, Chengda and Zhao, Chenggang and Deng, Chengqi and Zhang, Chenyu and Ruan, Chong and others},
journal={arXiv preprint arXiv:2412.19437},
year={2024}
}

@article{zhang2025igniting,
  title={Igniting language intelligence: The hitchhiker’s guide from chain-of-thought reasoning to language agents},
  author={Zhang, Zhuosheng and Yao, Yao and Zhang, Aston and Tang, Xiangru and Ma, Xinbei and He, Zhiwei and Wang, Yiming and Gerstein, Mark and Wang, Rui and Liu, Gongshen and others},
  journal={ACM Computing Surveys},
  volume={57},
  number={8},
  pages={1--39},
  year={2025},
  publisher={ACM New York, NY}
}

\section{Biography Section}

\vspace{-21pt}
\begin{IEEEbiography}[{\includegraphics[width=1in,height=1.25in,clip,keepaspectratio]{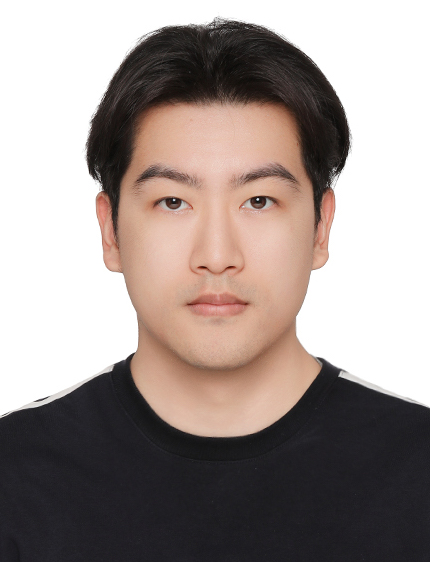}}]{Yunzhe Xu}
received the bachelor's degree in software engineering from Harbin Institute of Technology in 2022. He is currently pursuing the Ph.D. degree in computer science and technology with Shanghai Jiao Tong University. His research interests include embodied navigation system, robotic learning and large language model agents.
\end{IEEEbiography}

\vspace{-21pt}
\begin{IEEEbiography}[{\includegraphics[width=1in,height=1.25in,clip,keepaspectratio]{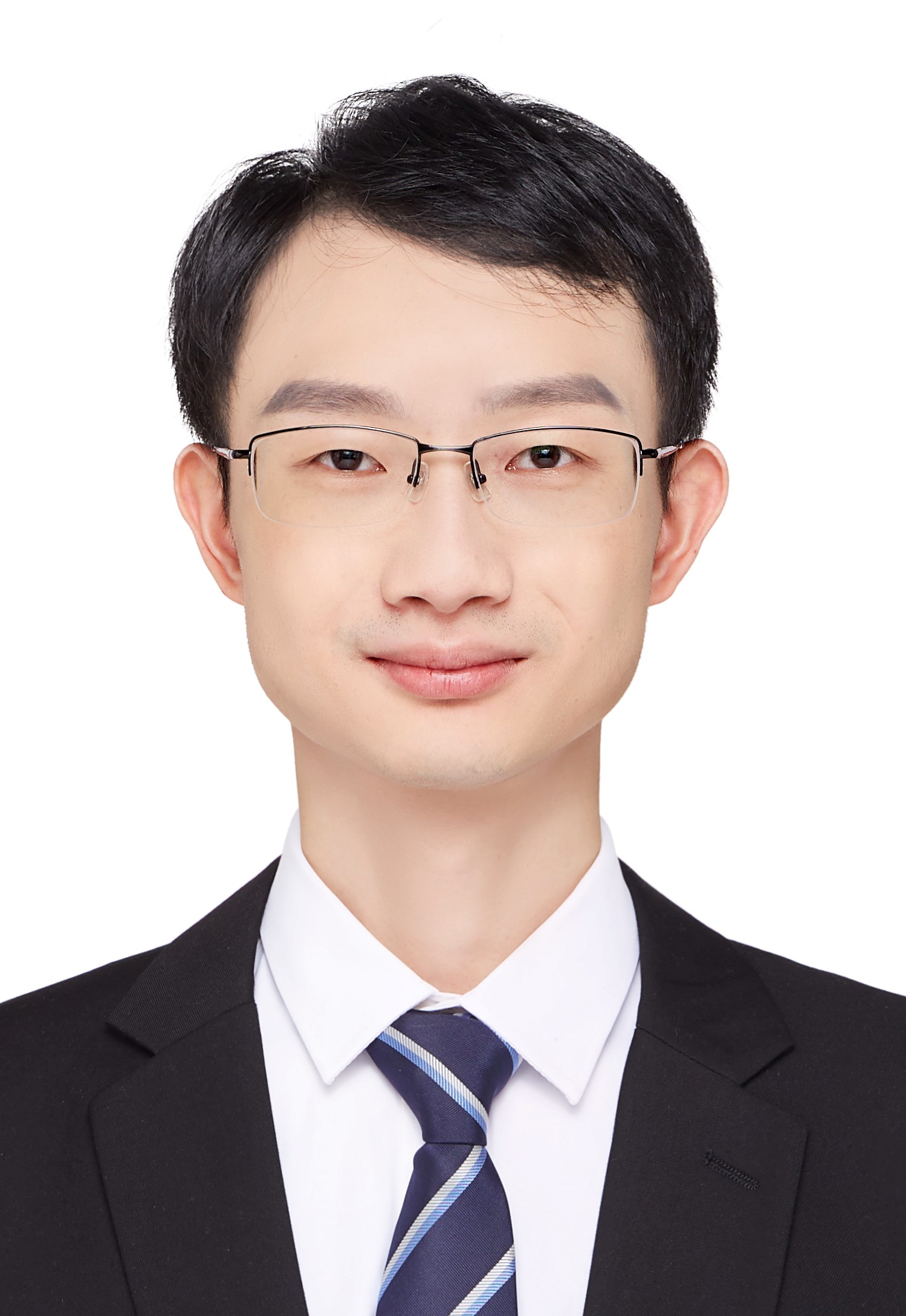}}]{Zhuosheng Zhang} is currently a tenure-track assistant professor at School of Computer Science, Shanghai Jiao Tong University. He received his Bachelor's degree from Wuhan University in 2016, and his M.S. and Ph.D. degrees from Shanghai Jiao Tong University in 2020 and 2023, respectively. He was a research intern at Amazon AWS, Microsoft Research, Langboat Technology, NICT (Japan), and IBM. His research interests include natural language processing, large language models, and language agents. He has published about 80 research papers in leading journals and conferences, such as TPAMI, TNNLS, TASLP, ICLR, ICML, ACL, AAAI, EMNLP, and COLING. He was the recipient of the WAIC 2024 Youth Outstanding Paper Award, WAIC 2024 YunFan Award, and the Global Top 100 Chinese Rising Stars in Artificial Intelligence. He serves as an action editor for ACL Rolling Review and standing reviewer for TACL. He served as the area chair or senior program committee for international conferences such as NeurIPS, AAAI, IJCAI, ACL, EMNLP, and COLING.
\end{IEEEbiography}

\vspace{-21pt}
\begin{IEEEbiography}[{\includegraphics[width=1in,height=1.25in,clip,keepaspectratio]{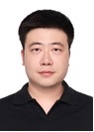}}]{Zhe Liu}
received the Ph.D. degree in control technology and control engineering from Shanghai Jiao Tong University, Shanghai, China, in 2016. From 2017 to 2020, he was a Post-Doctoral Fellow with the Department of Mechanical and Automation Engineering, The Chinese University of Hong Kong, Hong Kong. From 2020 to 2022, he was a Research Associate with the Department of Computer Science and Technology, University of Cambridge, Cambridge, U.K. From 2022 to 2025, he has been an Associate Professor with the AI institute, Shanghai Jiao Tong University, where he is currently an Associate Professor with the Department of Automation. His current research interests include multi-robot cooperation and autonomous driving system.
\end{IEEEbiography}

\vfill

\end{document}